\documentclass[sn-basic]{sn-jnl_mod}

\hypersetup{pdftitle={Hydra}}

\newcommand{\hydra}{\textsc{Hydra}}
\newcommand{\rocket}{\textsc{Rocket}}
\newcommand{\minirocket}{\textsc{MiniRocket}}
\newcommand{\multirocket}{\textsc{MultiRocket}}

\begin{document}

\title[{\hydra}]{\vspace{-20mm}{\hydra}}
\subtitle{Competing convolutional kernels for fast and accurate time series classification}

\author*{\fnm{Angus} \sur{Dempster}}\email{angus.dempster1@monash.edu}
\author{\fnm{Daniel F.} \sur{Schmidt}}\nomail
\author{\fnm{Geoffrey I.} \sur{Webb}}\nomail

\affil{\orgname{Monash University}, \orgaddress{\city{Melbourne}, \country{Australia}}}

\jyear{2022}

\abstract{We demonstrate a simple connection between dictionary methods for time series classification, which involve extracting and counting symbolic patterns in time series, and methods based on transforming input time series using convolutional kernels, namely {\rocket} and its variants.  We show that by adjusting a single hyperparameter it is possible to move by degrees between models resembling dictionary methods and models resembling {\rocket}.  We present {\hydra}, a simple, fast, and accurate dictionary method for time series classification using competing convolutional kernels, combining key aspects of both {\rocket} and conventional dictionary methods.  {\hydra} is faster and more accurate than the most accurate existing dictionary methods, and can be combined with {\rocket} and its variants to further improve the accuracy of these methods.}


\maketitle

\vspace{-5mm}
\section{Introduction} \label{section-introduction}

\begin{figure}
\centering
\includegraphics[width=1.0\linewidth]{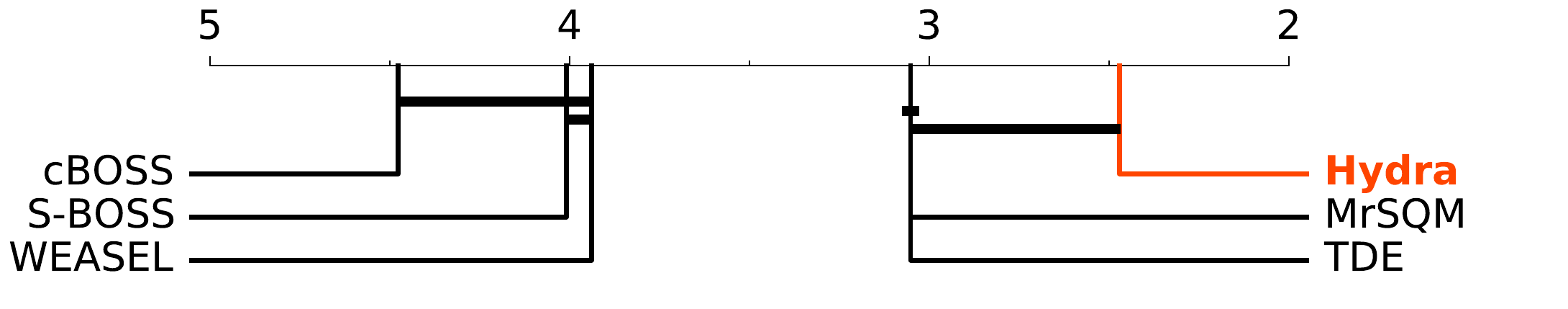}
\caption{Mean rank of {\hydra} in terms of accuracy versus several prominent dictionary methods over 30 resamples of 106 datasets from the UCR archive.  Lower rank indicates higher accuracy.  {\hydra} is more accurate than the most accurate existing dictionary methods.}
\label{fig-rank-ucr-dict}
\end{figure}

\begin{figure}
\centering
\includegraphics[width=0.5\linewidth]{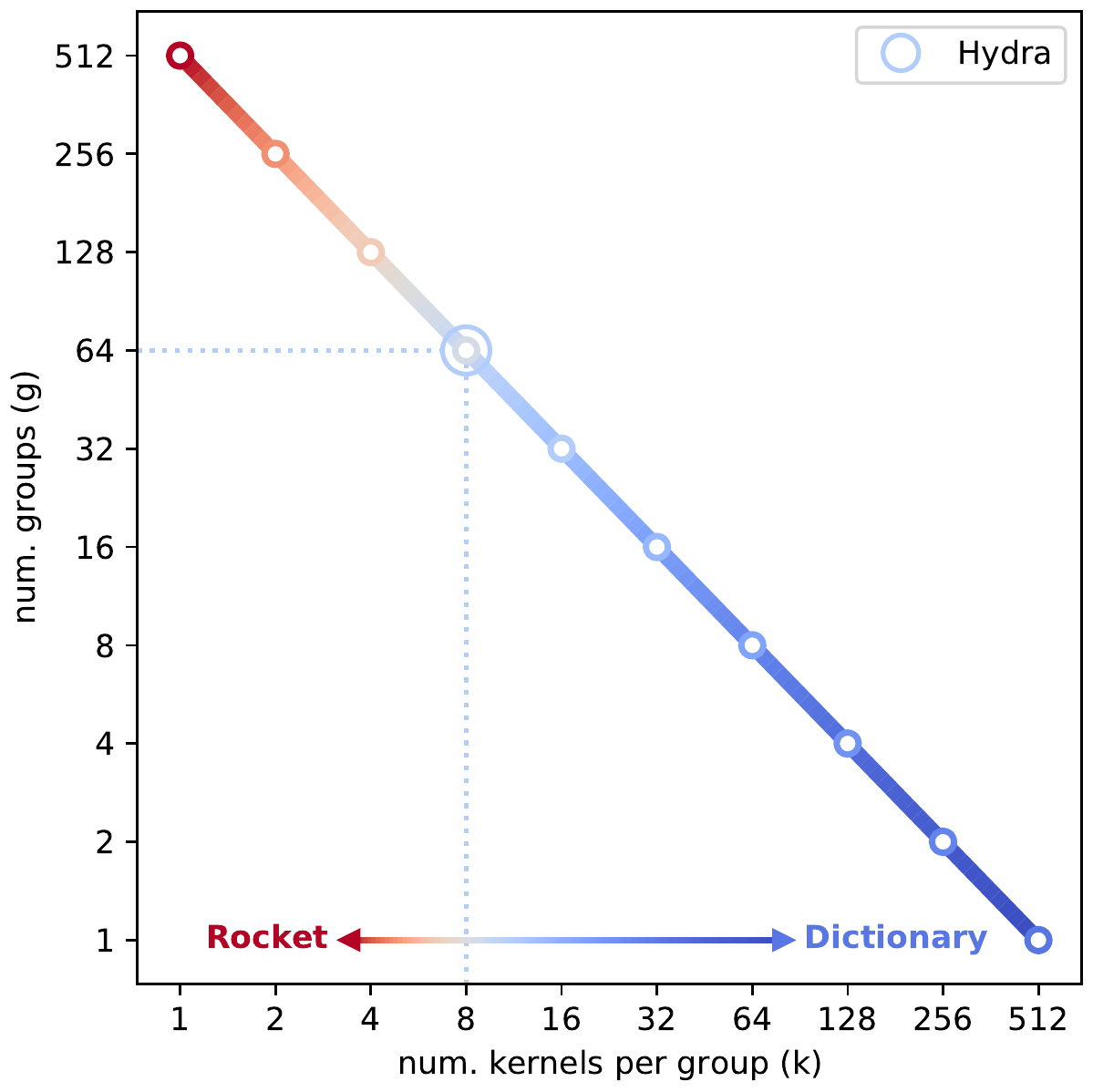}
\caption{The number of groups ($g$) or, equivalently, the number of kernels per group ($k$), controls the resemblance of {\hydra} to dictionary methods or {\rocket}.  By default, {\hydra} uses 64 groups with 8 kernels per group.}
\label{fig-diagram-spectrum}
\end{figure}

Dictionary methods and {\rocket} \citep{dempster_etal_2020} represent two seemingly quite different approaches to time series classification.  Dictionary methods involve extracting and then counting symbolic patterns in time series.  {\rocket} and its variants transform time series using convolutional kernels, and could be seen as having more in common with convolutional neural networks.

We demonstrate a simple connection between dictionary methods and {\rocket}.  We show that by adjusting a single hyperparameter it is possible to move by degrees between models more closely resembling dictionary methods and models more closely resembling {\rocket}.  This provides the basis for performing fast and accurate dictionary-like classification using convolutional kernels.  We present {\hydra} (for \textbf{HY}brid \textbf{D}ictionary--\textsc{\textbf{R}ocket} \textbf{A}rchitecture), a simple, fast, and accurate dictionary method for time series classification using competing convolutional kernels, incorporating aspects of both {\rocket} and conventional dictionary methods.

{\hydra} incorporates two key aspects of dictionary methods: (1) forming groups (dictionaries) of patterns which approximate the input time series; and (2) using the counts of these patterns to perform classification.  However, unlike typical dictionary methods, {\hydra} uses random patterns represented by random convolutional kernels.

Like {\rocket}, {\hydra} transforms input time series using random convolutional kernels.  However, unlike {\rocket}: (1) the kernels are arranged into groups; and (2) {\hydra} counts the kernels in each group representing the closest match with the input at each timepoint, i.e., treating each kernel as a pattern in a dictionary, and forcing the kernels in each group to `compete' in order to be counted at each timepoint (see Figure \ref{fig-diagram-overview}).

While superficially simple, the organisation of the kernels into groups is what allows {\hydra} to perform dictionary-like classification using random patterns with high accuracy.  The number of groups ($g$) or, equivalently, the number of kernels per group ($k$),\footnote{For a fixed value of $k \times g$, increasing $k$ implies decreasing $g$, and vice versa.} controls the extent to which {\hydra} more closely resembles dictionary methods or more closely resembles {\rocket}, and has a decisive influence on the accuracy of the method: see Figure \ref{fig-diagram-spectrum}.  With multiple kernels per group, {\hydra} more closely resembles conventional dictionary methods.  As the number of kernels per group approaches one ($k \rightarrow 1$), {\hydra} more closely resembles {\rocket}.  In fact, where $k = 1$ ($g \gg 1$), {\hydra} essentially \textit{is} {\rocket}, with some qualifications.  However, the most {\rocket}-like variant is \textit{not} the most accurate variant of {\hydra}.

The use of random patterns in the form of random convolutional kernels has two key advantages: (1) it is computationally efficient; and (2) it produces high classification accuracy.  Figure \ref{fig-rank-ucr-dict} shows the mean rank of {\hydra} versus several prominent dictionary methods---cBOSS, S-BOSS, WEASEL, TDE, and MrSQM---over 30 resamples of the 106 datasets from the UCR archive for which benchmark results are available for all relevant methods.  We rank the accuracy of each method on each dataset, and take the mean rank over all 106 datasets.  Lower mean rank corresponds to higher accuracy.  Methods for which the pairwise difference in accuracy is not statistically significant, per a Wilcoxon signed-rank test with Holm correction, are connected with a black line \citep[see][]{demsar_2006,garcia_and_herrera_2008,benavoli_etal_2016}.

{\hydra} is faster and more accurate than the most accurate existing dictionary methods, TDE and MrSQM, and is competitive in terms of accuracy with several of the most accurate current methods for time series classification: see Section \ref{subsection-experiments-ucr}.  {\hydra} can train and test on this subset of 106 datasets in approximately 28 minutes using a single CPU core, as compared to approximately 1 hour 41 minutes for MrSQM, 4 hours for cBOSS, 22 hours for TDE, more than 24 hours for WEASEL, and more than 6 days for S-BOSS.  Compute times for {\hydra} and MrSQM are averages over 30 resamples, run on a cluster using Intel Xeon E5-2680 and Xeon Gold 6150 CPUs, restricted to a single CPU core per dataset per resample.  Compute times for cBOSS, TDE, S-BOSS, and WEASEL are taken from \citet{middlehurst_etal_2020a}.

The rest of this paper is structured as follows.  In Section \ref{section-related-work}, we present relevant related work.  In Section \ref{section-method}, we explain {\hydra} in detail.  In Section~\ref{section-experiments}, we present experimental results including results for a number of larger datasets, and a sensitivity analysis for key hyperparameters.

\section{Related Work} \label{section-related-work}

\subsection{Dictionary Methods} \label{subsection-related-dict}

Dictionary (or `bag of words') methods represent a prominent approach to time series classification.  BOSS was identified in \citet{bagnall_etal_2017} (the `bake off' paper) as one of the three most accurate methods for time series classification on the datasets in the UCR archive \citep{dau_etal_2019}, and was included in the original HIVE-COTE ensemble, then the most accurate method for time series classification on the datasets in the UCR archive \citep{lines_etal_2018}.  The most accurate current dictionary methods, TDE \citep{middlehurst_etal_2020a} and MrSQM \citep{lenguyen_and_ifrim_2022}, are competitive with several of the most accurate current methods for time series classification on the datasets in the UCR archive.  TDE is one of the four components of HIVE-COTE 2 (HC2), currently the most accurate method for time series classification on the datasets in the UCR archive \citep{middlehurst_etal_2021}.

Most dictionary methods work in broadly the same way, i.e., by passing a sliding window over each time series, smoothing or approximating the values in each window, and assigning the resulting values to letters from a symbolic alphabet \citep{large_etal_2019}.  The counts of the resulting patterns are used as the basis for classification.  BOSS and several other methods use Symbolic Fourier Approximation (SFA) \citep{schafer_2015}, which involves:
\begin{itemize}
  \item passing a sliding window over the input time series;
  \item applying the Fourier transform to the values in the window, dropping the high-frequency coefficients (in effect, smoothing the values in the window using a low-pass filter); and
  \item assigning the remaining values to one of four letters to form words.
\end{itemize}

The counts of the resulting words are used to perform classification with a 1-nearest neighbour (1NN) classifier.  BOSS is a large ensemble of such classifiers using different hyperparameter configurations.

The resulting feature space is typically very large and very sparse \citep[see][]{schafer_and_leser_2016,large_etal_2019,lenguyen_and_ifrim_2022}, and the resulting patterns represent a high degree of approximation, as the input is both smoothed and quantised to a very small set of discrete values.  In addition, for methods using SFA or a variation thereof, the patterns are formed over values in the frequency domain, rather than the original input.

Despite the broad similarities between many dictionary methods, different methods can produce very different results due to differences in the way that the input is approximated or quantised \citep{bagnall_etal_2017,large_etal_2019}.  The most prominent dictionary methods are BOSS \citep{schafer_2015}, cBOSS \citep{middlehurst_etal_2019}, S-BOSS \citep{large_etal_2019}, WEASEL \citep{schafer_and_leser_2016} and, more recently, TDE \citep{middlehurst_etal_2020a} and MrSQM \citep{lenguyen_and_ifrim_2022}.

In contrast to BOSS, cBOSS randomly selects hyperparameter combinations for its ensemble, sets an upper limit on ensemble size, and uses a different ensemble weighting.  cBOSS is considerably faster than BOSS with approximately the same accuracy on the datasets in the UCR archive.

S-BOSS adds temporal information by recursively dividing the input time series into subseries, and forming dictionaries over the subseries.  S-BOSS also uses a different distance measure for the 1NN classifiers.  S-BOSS is more accurate than BOSS on the datasets in the UCR archive, but at considerable computational expense.

WEASEL selects Fourier coefficients using a statistical test, performs quantisation based on information gain, and uses a chi-squared test to perform feature selection.  WEASEL uses the resulting features to train a logistic regression model.  WEASEL is more accurate than BOSS or cBOSS on the datasets in the UCR archive.

TDE incorporates the ensembling method from cBOSS, the temporal information and distance measure from S-BOSS, and the quantisation method from WEASEL.  TDE is currently one of the two the most accurate dictionary methods on the datasets in the UCR archive and, as noted above, is one of the four components of HC2.

MrSQM builds on an earlier method, MrSEQL \citep{lenguyen_etal_2019}.  MrSQM uses a combination of random feature selection and feature selection via a chi-squared test, and uses the resulting features to train a logistic regression model.  MrSQM is both significantly more accurate, and an order of magnitude faster, than MrSEQL.  Along with TDE, MrSQM is one of the two most accurate dictionary methods on the datasets in the UCR archive.

\subsection{{\rocket}, {\minirocket}, and {\multirocket}}

The {\rocket} `family' of methods---namely {\rocket} and its variants {\minirocket} \citep{dempster_etal_2021} and {\multirocket} \citep{tan_etal_2022}---represent a seemingly quite different approach to time series classification.

{\rocket} transforms input time series using a large number of random convolutional kernels (by default, $10{,}000$), and uses the transformed features to train a linear classifier.  {\rocket} uses kernels with lengths selected randomly from $\{7, 9, 11\}$, weights drawn from $\mathcal{N}(0, 1)$, biases drawn from $\mathcal{U}(-1, 1)$, random dilation (on an exponential scale), and random padding.  {\rocket} applies both global max pooling, and `proportion of positive values' (PPV) pooling to the convolution output.  The resulting features are used to train a ridge regression classifier or logistic regression (for larger datasets).  The two most important aspects of {\rocket} in terms of accuracy are the use of dilation and PPV pooling.  {\rocket} achieves state-of-the-art accuracy with a fraction of the computational expense of other methods of comparable accuracy, with the exception of {\minirocket} and {\multirocket}.  Along with TDE, an ensemble of {\rocket} models, known as Arsenal, is one of the four components of HC2 \citep{middlehurst_etal_2021}.

{\minirocket} makes several key changes to {\rocket} in order to remove randomness and significantly speed up the transform.  {\minirocket} uses a fixed kernel length of 9, a small, fixed set of 84 kernels, bias values drawn from the convolution output, a fixed set of dilation values (fixed relative to the length of the input time series), and only produces PPV features.  {\minirocket} uses the same classifiers as {\rocket}.  {\minirocket} is significantly faster than any other method of comparable accuracy, including {\rocket}, and demonstrates that it is possible to achieve essentially the same accuracy as {\rocket} using a fixed set of nonrandom kernels.

{\multirocket} represents a further extension of {\rocket}/{\minirocket}, adding three additional pooling operations (in addition to PPV), as well as transforming both the original time series and the first-order difference.  {\multirocket} uses the same kernels and classifiers as {\minirocket}.  {\multirocket} is currently the next-most-accurate method for time series classification after HC2 on the datasets in the UCR archive.  {\multirocket} achieves broadly similar accuracy to HC2 while being several orders of magnitude faster.

While seemingly very different, we show that there is, in fact, a simple connection between {\rocket} and dictionary methods, and that this connection provides the basis for performing fast and accurate dictionary-like classification using convolutional kernels.

\subsection{Other State of the Art}

There has been significant progress in terms of accuracy in time series classification since \citet{bagnall_etal_2017}.  In addition to {\rocket} and its variants, the most accurate current methods on the datasets in the UCR archive include TDE, MrSQM, DrCIF \citep{middlehurst_etal_2020b,middlehurst_etal_2021}, InceptionTime \citep{ismailfawaz_etal_2020}, TS-CHIEF \citep{shifaz_etal_2020}, and HC2 \citep[see][]{bagnall_etal_2020,middlehurst_etal_2021}.

DrCIF builds on an earlier method, TSF, to extract multiple features, including `catch22' features \citep{lubba_etal_2019}, from random intervals.  DrCIF takes intervals from the input time series, the first-order difference, and a periodogram.  Like TDE and Arsenal, DrCIF is one of the components of HC2.

InceptionTime is an ensemble of deep convolutional neural networks based on the Inception architecture, and is currently the most accurate convolutional neural network model for time series classification on the UCR archive.

TS-CHIEF is an extension of an earlier method, ProximityForest \citep{lucas_etal_2019}, and is an ensemble of decision trees using distance measures, intervals, and spectral splitting criteria.

HC2 supersedes earlier variants of HIVE-COTE \citep{lines_etal_2018,bagnall_etal_2020}, and is an ensemble comprising TDE, DrCIF, Shapelet Transform, and Arsenal.  HC2 is currently the most accurate method for time series classification on the datasets in the UCR archive.

These methods, while highly accurate, are all burdened by high computational complexity.  DrCIF takes almost 2 days to train and test on 112 datasets in the UCR archive, TDE more than 3 days, InceptionTime more than 3 days (using GPUs), HC2 approximately 2 weeks, and TS-CHIEF several weeks \citep{middlehurst_etal_2021}.  MrSQM is faster, taking approximately 3 hours to train and test on the same datasets using broadly comparable hardware.  In comparison, {\hydra} completes the same task in approximately 36 minutes.

\section{Method} \label{section-method}

\subsection{Overview}

\begin{figure}
\centering
\includegraphics[width=1.0\linewidth]{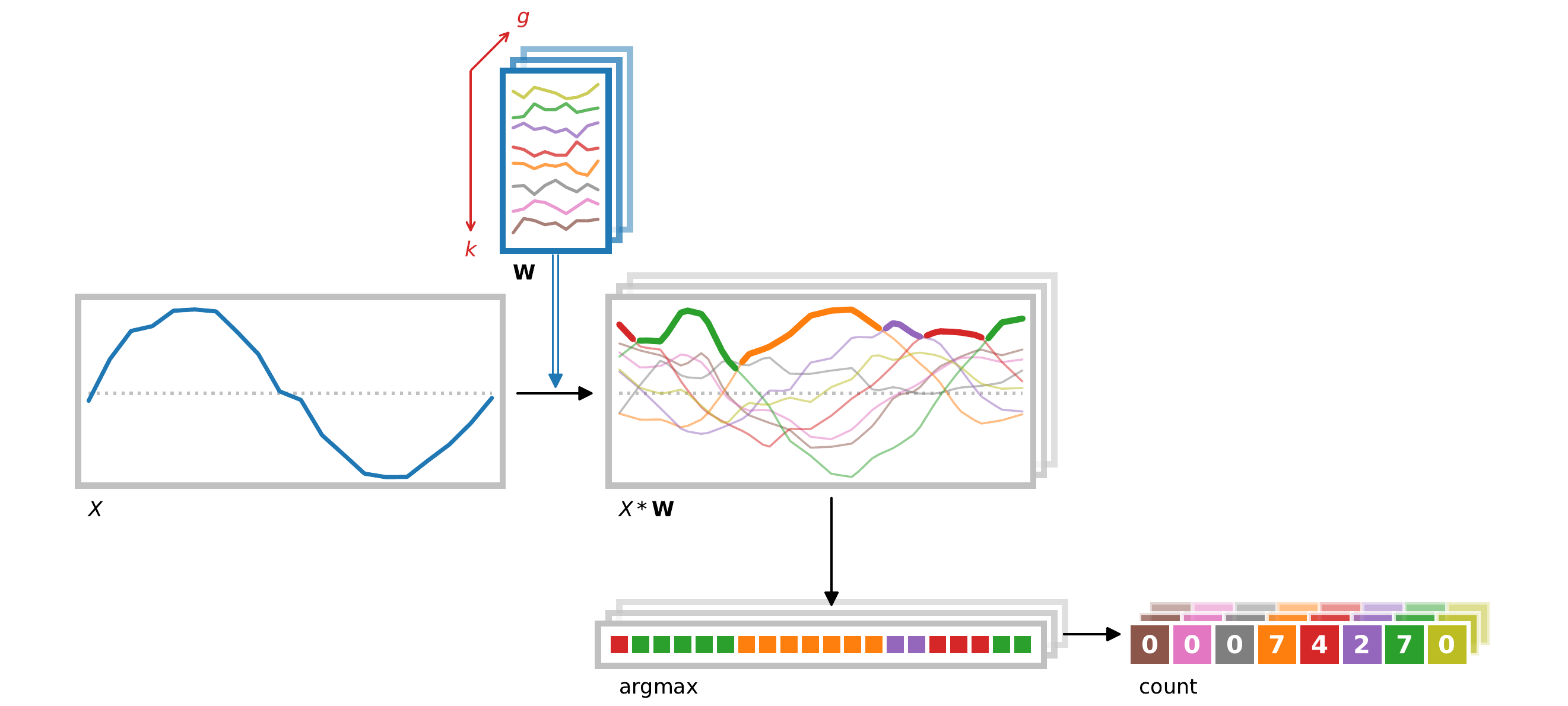}
\caption{{\hydra} convolves each input time series with a set of random convolutional kernels, organised into $g$ groups with $k$ kernels per group, and at each timepoint counts the kernels representing the closest match with the input time series for each group.}
\label{fig-diagram-overview}
\end{figure}

{\hydra} is a dictionary method which uses convolutional kernels, incorporating aspects of both {\rocket} and conventional dictionary methods.

{\hydra} involves transforming the input time series using a set of random convolutional kernels, arranged into $g$ groups with $k$ kernels per group, and then at each timepoint counting the kernels representing the closest match with the input time series for each group: see Figure \ref{fig-diagram-overview}.  The counts are then used to train a linear classifier.  (Note that there are $g$ groups \textit{per dilation}.)

Like other dictionary methods, {\hydra} uses patterns which approximate the input, and produces features which represent the counts of those patterns.  However, unlike typical dictionary methods, {\hydra} uses random patterns, represented by random convolutional kernels.

While the use of random patterns distinguishes {\hydra} from typical dictionary methods, the difference is not as radical as it might first appear.  The patterns extracted by typical dictionary methods represent a considerable degree of approximation of the input time series: see Section \ref{subsection-related-dict}.  Although they represent a different form of approximation (randomisation, rather than smoothing and quantisation), the patterns identified and counted by {\hydra} are not necessarily `more approximate' than the patterns used in typical dictionary methods.

Like {\rocket}, {\hydra} transforms the input time series using random convolutional kernels.  However, unlike {\rocket} and its variants: (1) the kernels are organised into groups; and (2) {\hydra} counts the kernels in each group representing the closest match with the input at each timepoint.  In effect, {\hydra} treats each kernel as a pattern in a dictionary, and treats each group as a dictionary.  In a sense, the kernels in each group are forced to ‘compete’ in order to be counted at each timepoint.

The organisation of the kernels into groups, although seemingly simple, is what allows us to move between models more closely resembling dictionary methods and models more closely resembling {\rocket}, and has a decisive influence on the accuracy of the method.  Further, the move away from producing PPV features represents a major departure from the {\rocket} paradigm.  PPV pooling is a defining characteristic of the {\rocket} family of methods, and one of the most important factors allowing these methods to achieve high accuracy.

\begin{samepage}
\noindent The key hyperparameters for {\hydra} are:
\begin{itemize}
  \item \textit{the characteristics of the kernels}---for simplicity, {\hydra} largely inherits the characteristics of the kernels from {\rocket} and its variants (Section \ref{subsection-method-kernels});
  \item \textit{the way in which the kernels are counted}---by default, at each timepoint {\hydra} counts both the kernel with the maximum response and the kernel with the minimum response, using both `hard' and `soft' counting, as explained below (Section \ref{subsection-method-counting});
  \item \textit{the number of groups and the number of kernels per group}---by default, {\hydra} uses 64 groups ($g = 64$) with 8 kernels per group ($k = 8$), for a total of $k \times g = 512$ kernels per dilation (Section \ref{subsection-method-k-g}); and
  \item \textit{whether not to include the first-order difference}---by default, {\hydra} acts on both the original time series as well as the first-order difference (Section~\ref{subsection-method-diff}).
\end{itemize}
\end{samepage}

In setting the values of these hyperparameters, we have restricted ourselves to the same subset of 40 `development' datasets (default training/test splits) from the UCR archive per \citet{dempster_etal_2020,dempster_etal_2021}, in order to avoid overfitting the entire archive.

\begin{lstlisting}[label={listing-pseudocode},escapechar=@,caption={PyTorch-like pseudocode for {\hydra}.  The transform involves convolving the input with the kernels, rearranging the output into $g$ groups, performing an argmax (and/or max) operation, and incrementing the counts for the relevant kernels.},float]
# X : n input time series
# W : g * k kernels
def transform(X, W):
    indices = conv1d(X, W).reshape(n, g, k, -1).argmax(2)
    return zeros(n, g, k).scatter_add_(-1, indices, ones_like(indices))
\end{lstlisting}

{\hydra} is conducive to a very simple implementation: see the pseudocode in Listing \ref{listing-pseudocode}.  We implement {\hydra} using PyTorch \citep{paszke_etal_2019}.  We use the ridge regression classifier from scikit-learn \citep{pedregosa_etal_2011}, and logistic regression implemented using PyTorch.  Our code and full results will be made available at \url{https://github.com/angus924/hydra}.

\subsection{Kernels} \label{subsection-method-kernels}

{\hydra} inherits major kernel characteristics from {\rocket} and its variants:
\begin{itemize}
  \item a kernel length of 9 (per {\minirocket}/{\multirocket});
  \item weights drawn from $\mathcal{N}(0, 1)$ (per {\rocket}); and
  \item exponential dilation.
\end{itemize}

Bias values, closely linked to the PPV features produced by {\rocket} and its variants, are not used at all in {\hydra}.  As such, we do not have the same flexibility to `spread' features over different dilations (e.g., by assigning more bias values to smaller or larger dilations), or to increase or decrease the total number of features (e.g., by increasing or decreasing the total number of bias values).  Accordingly, {\hydra} uses a simplified set of dilations in the range $\{2^0, 2^1, 2^2, ...\}$ (such that the maximum effective length of a kernel including dilation is the length of the input time series), and does not use a fixed feature count (c.f., approx. $10{,}000$ for {\minirocket}, $20{,}000$ for {\rocket}, and approx. $50{,}000$ for {\multirocket}).  Additionally, for simplicity, {\hydra} always uses padding (zeros are added to the start and end of each time series such that the convolution operation begins with the middle element of the kernel centred on the first element of the series and ends with the middle element of the kernel centred on the last element of the series).  Using a fixed kernel length, simplified dilation, and consistent padding, greatly simplifies the organisation of kernels into groups.  Always padding has the additional advantage of allowing the transform to work by default with variable-length time series.

As noted above, there are $g$ groups \textit{per dilation}.  In this sense, dilation acts as a kind of super grouping of kernels.  Accordingly, the total number of kernels ($k \times g \times d$, where $d$ is the number of dilations) will grow logarithmically with time series length and, for very long time series, it may be desirable to, e.g., subsample the dilation values, use a smaller number of groups (per dilation), and/or downsample the input.  However, we have not used any such restrictions in any of the experiments presented in this paper.

We normalise the kernel weights by subtracting the mean and dividing by the sum of the absolute values.  This is to ensure that one kernel is not counted over another kernel because of a spurious difference in the mean or magnitude of the kernel weights.  (The scale of the weights is unimportant, provided that the weights for the kernels in each group are on the same scale.)

It is possible to perform the transform using a smaller set of kernels, e.g., the {\minirocket} kernels, and to form the groups by resampling the convolution output, i.e., such that the convolution output for each kernel may appear in more than one group.  This could improve speed and scalability, and potentially provide the basis for a deterministic transform.  However, initial experimentation suggests that accuracy decreases as the number of unique kernels decreases.

As noted above, {\minirocket} and {\multirocket} achieve high accuracy without using random kernels.  This suggests that it is not necessarily the `randomness' of the kernels used in {\hydra} which is important.  It is possible that a larger set of {\minirocket}-like kernels would achieve similar accuracy to the random kernels used in {\hydra}.  However, it is not clear whether a fixed grouping or predefined resampling would be effective, or whether the resampling procedure should remain stochastic, even for nonrandom kernels.  We leave the exploration of these questions for future work.

\subsection{Method of Counting} \label{subsection-method-counting}

\subsubsection{Maximum vs Minimum Response}

By default, at each timepoint {\hydra} counts both the kernel with the maximum response (output value), i.e., the kernel representing the closest match with the input time series, and the kernel with the minimum response.

Unlike {\rocket} and its variants, where a kernel is counted in {\hydra}, the information for the other kernels at that timepoint is discarded.  Counting the kernel with the minimum response, in addition to the kernel with the maximum response, preserves more of the information in the convolution output without requiring additional work in terms of the transform, and improves accuracy: see Section \ref{subsection-experiments-sensitivity}.  Note, however, that where $k = 1$, counting both the minimum response and the maximum response is redundant: they are the same.

The kernel with the minimum response does not represent the `poorest match' to the input time series, but rather an alternative closest match.  (The `poorest match' would be the kernel with the lowest-magnitude response.)  Counting both the kernel with the maximum response and the kernel with the minimum response is equivalent to looking for the closest match twice: once with the given set of kernels, $\boldsymbol{W}$, and once with the `inverted' set of kernels, $-\boldsymbol{W}$.  (As the kernels are random, the sign of the kernels is arbitrary.)

\subsubsection{Hard (argmax) vs Soft (max) Counting}

{\hydra} uses two different forms of counting:

\begin{itemize}
  \item \textit{`hard' counting}---incrementing the count of the kernel with the maximum (and/or minimum) response at each timepoint; and
  \item \textit{`soft' counting}---accumulating the value of the maximum (and/or minimum) response for the kernel with the maximum (and/or minimum) response at each timepoint.
\end{itemize}

Hard counting involves an argmax (or argmin) operation over the channels in the convolution output for each group (one channel per kernel), returning the index of the kernel with the maximum (or minimum) response at each timepoint.  This is then used to increment the counts for the relevant kernels.

Soft counting involves a max (or min) operation over the same channels, returning the maximum (or minimum) response at each timepoint.  This is then accumulated for the relevant kernels.

Although superficially different, both hard and soft counting capture broadly similar information.  Soft counts can be expected to be roughly proportional to hard counts, and vice versa.  Where a kernel predominates, it will have a higher relative count, and the sum of the response (where that kernel is counted) is also likely to be relatively large.

By default, {\hydra} uses both hard counting and soft counting: soft counting for the maximum response, and hard counting for the minimum response.  (The allocation of soft counting to the maximum response and hard counting to the minimum response is essentially arbitrary.)  Using both soft and hard counting is more accurate than using either alone: see Section \ref{subsubsection-counting}.

\subsubsection{Clipping}

By default, {\hydra} counts the kernel with the maximum response whether or not the maximum response is positive (and, likewise, counts the kernel with the minimum response whether or not the minimum response is negative).

It is possible to `clip' the values in the convolution output, such that the kernel with the maximum response is only counted when the maximum response is positive (and, likewise, the kernel with the minimum response is only counted when the minimum response is negative).  This is essentially equivalent to passing the convolution output through a ReLU function.

Without clipping, the features produced by the most {\rocket}-like variant (i.e., $k = 1$) are uninformative:
\begin{itemize}
  \item hard counting implies counting every timepoint for every kernel, such that all the features are the same, i.e., a value representing the length of the input time series; and
  \item soft counting produces features which are approximately zero, as positive and negative values in the convolution output will `cancel each other out'.
\end{itemize}

With clipping, the features produced where $k = 1$ are equivalent to PPV (hard counting), and global average pooling (soft counting): see Section \ref{subsubsection-features}.  However, clipping is of little practical significance.  While it improves accuracy for $k = 1$, it has little or no effect for $k \geq 2$, and has no effect on the optimal values of $g$ and $k$.

\subsection{Number of Groups ($g$) \& Kernels Per Group ($k$)} \label{subsection-method-k-g}

The number of groups ($g$) or, equivalently, the number of kernels per group ($k$), controls the extent to which {\hydra} more closely resembles dictionary methods or more closely resembles {\rocket} in two senses:

\begin{itemize}
  \item \textit{the features produced by the transform}---whether the features are more like the features produced by dictionary methods or more like the features produced by {\rocket}; and
  \item \textit{the level of approximation provided by the random kernels}---whether the patterns identified and counted by {\hydra} are more, or less, representative of the input time series.
\end{itemize}

\subsubsection{Features} \label{subsubsection-features}

With multiple kernels per group ($k >1$), {\hydra} produces counts like typical dictionary methods.  However, as the number of kernels per group decreases, in particular, as the number of kernels per group approaches one ($k \rightarrow 1$), the features produced by the transform undergo a qualitative change.

Where $k = 1$, provided that at each timepoint a kernel is only counted if the maximum response is positive (or the minimum response is negative), {\hydra} produces PPV and/or global average pooling features:

\begin{itemize}
  \item hard counting involves incrementing the count for every kernel wherever the convolution output is positive, i.e., PPV; and
  \item soft counting involves accumulating the output values for each kernel wherever the convolution output is positive, i.e., global average pooling.
\end{itemize}

In other words, where $k = 1$, the features produced by {\hydra} no longer represent counts, but instead represent the average response of every kernel across all timepoints in the form of PPV and/or global average pooling.  In this sense, $k = 1$ represents the most {\rocket}-like variant of {\hydra}.

However, as noted above, {\hydra} only produces PPV and/or global average pooling features where $k = 1$, and then only where the values in the convolution output are `clipped'.  Additionally, the most {\rocket}-like variant of {\hydra} is \textit{not} the most accurate variant of {\hydra}, whether or not clipping is used.  This may be, in part, because the effectiveness of PPV features is closely linked to the use of a large number of bias values and {\hydra}, in contrast to {\rocket} and its variants, does not use bias values at all.

\subsubsection{Approximation} \label{subsubsection-approximation}

\begin{figure}
\centering
\includegraphics[width=1.0\linewidth]{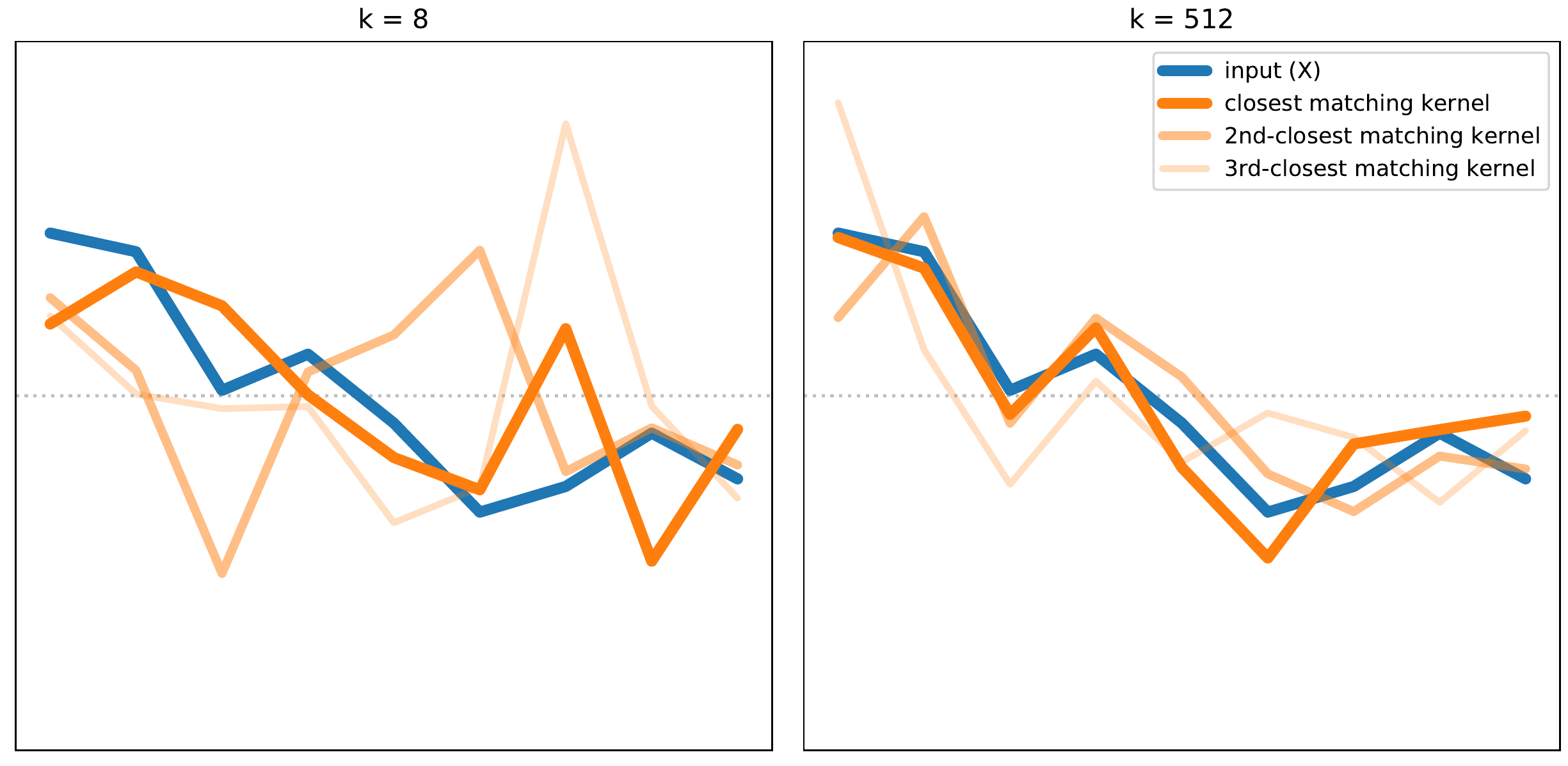}
\caption{A short time series segment is shown in blue.  The kernels representing the three closest matches to that segment are shown in orange, for a small ($k=8$) group, and for a large ($k=512$) group.  A small number of kernels per group (left) implies a high degree of approximation.  A large number of kernels per group (right) implies a low degree of approximation.}
\label{fig-diagram-approximation}
\end{figure}

A large number of kernels per group (i.e., a small number of groups) implies a low degree of approximation.  The larger the number of kernels per group, the more kernels are `competing' to have the maximum (or minimum) response at each timepoint, and the higher the probability that the kernels identified and counted by {\hydra} closely match the input time series: see Figure \ref{fig-diagram-approximation}.  (With an infinite number of kernels per group, the kernel with the maximum response at each timepoint would be expected to exactly match the input time series.)  In this sense, with a larger number of kernels per group, the patterns identified and counted by {\hydra} are more like patterns extracted from the input as in typical dictionary methods.

A small number of kernels per group implies a high degree of approximation.  The smaller the number of kernels per group, the fewer kernels are `competing', and the lower the probability that the kernels identified and counted by {\hydra} match the input time series.  As the number of kernels per group decreases, the patterns identified and counted by {\hydra} become increasingly approximate in the sense of becoming increasingly `random', i.e., unrelated to the input time series.

\subsection{First-Order Difference} \label{subsection-method-diff}

Like {\multirocket} and DrCIF, {\hydra} operates on both the original input time series and the first-order difference.  This is achieved by assigning half of the groups (i.e., by default, 32 of the 64 groups) to operate on the first-order difference.  This allows the transform to incorporate the first-order difference without introducing additional features or computation.  Adding the first-order difference increases the accuracy of every variant of {\hydra}: see Section~\ref{subsubsection-counting}.

\subsection{Classifier}

{\hydra} uses the same classifiers as {\rocket} and its variants, i.e., a ridge regression classifier or logistic regression (for larger datasets, i.e., where the number of training examples is greater than approx. {10{,}000}).  In order to limit peak memory usage, by default the transform is performed in batches.  As for {\rocket} and its variants, the {\hydra} transform can be naturally integrated into the minibatch updates for logistic regression trained using stochastic gradient descent or similar.

\section{Experiments} \label{section-experiments}

We evaluate {\hydra} on the datasets in the UCR archive (Section \ref{subsection-experiments-ucr}), demonstrating that {\hydra} is more accurate than the most accurate existing dictionary methods, TDE and MrSQM, and achieves similar accuracy to several of the most accurate current methods for time series classification.  We further demonstrate the accuracy of {\hydra} on a number of larger datasets (Section \ref{subsection-experiments-big}).  We also explore the effect of key hyperparameters in terms of the number of groups ($g$) and kernels per group ($k$), the way in which the kernels are counted, whether or not to include the first-order difference, and whether or not to `clip' the convolution output (Section \ref{subsection-experiments-sensitivity}).

\subsection{UCR Archive} \label{subsection-experiments-ucr}

\begin{figure}
\centering
\includegraphics[width=1.0\linewidth]{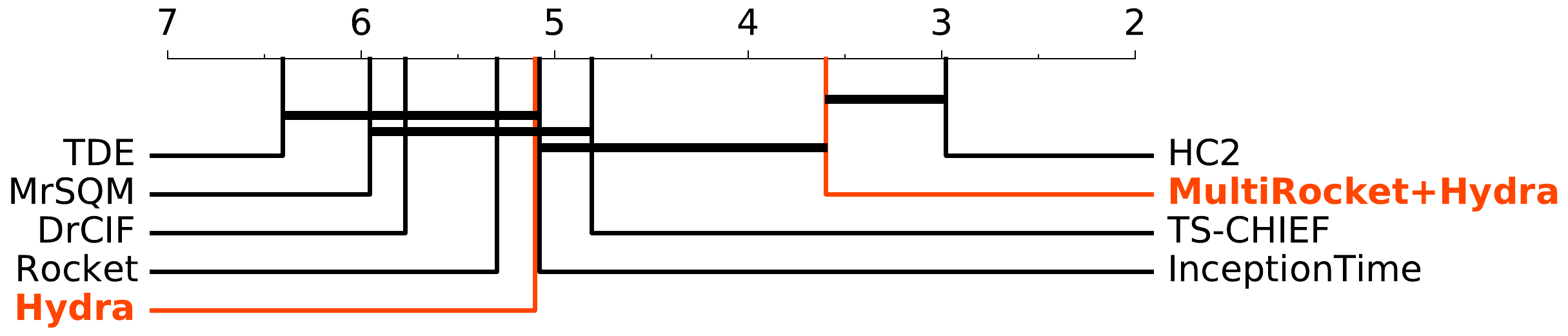}
\caption{Mean rank of {\hydra} and {\multirocket}+{\hydra} in terms of accuracy versus other SOTA methods over 30 resamples of 112 datasets from the UCR archive.}
\label{fig-rank-ucr-all}
\end{figure}

We evaluate {\hydra} on the datasets in the UCR archive \citep{dau_etal_2019}.  We compare {\hydra} against several prominent dictionary methods, and against the most accurate current methods for time series classification on the datasets in the UCR archive, namely, TDE, MrSQM, DrCIF, {\rocket} and its variants, InceptionTime, TS-CHIEF, and HC2.  For direct comparability with other published results, we evaluate {\hydra} on the same 30 resamples of 112 datasets from the UCR archive per \citet{middlehurst_etal_2020a,middlehurst_etal_2020b,middlehurst_etal_2021}.  The total compute time for {\hydra} on all 112 datasets, averaged over 30 resamples, is approximately 36 minutes using a single CPU core.

Figure \ref{fig-rank-ucr-dict} (page \pageref{fig-rank-ucr-dict}) shows the mean rank of {\hydra} versus several prominent dictionary methods, namely, cBOSS, S-BOSS, WEASEL, TDE, and MrSQM.  {\hydra} is faster and, on average, more accurate than all other dictionary methods, including TDE and MrSQM.  This ranking is performed on only 106 of the 112 datasets, as results for S-BOSS and WEASEL are unavailable for the six largest datasets due to computational constraints \citep[see][]{middlehurst_etal_2020a}.  We have obtained the results for MrSQM using the implementation available at \url{https://github.com/mlgig/mrsqm}, with the hyperparameter configuration as set out in \citet{lenguyen_and_ifrim_2022}.

Figure \ref{fig-rank-ucr-all} shows the mean rank of {\hydra} and {\multirocket}+{\hydra} versus the most accurate current methods for time series classification over 30 resamples of 112 datasets from the UCR archive.  {\hydra} is combined with {\multirocket} by simply concatenating the features produced by each transform.  On average, {\hydra} is more accurate (i.e., is more accurate on more than half the datasets) than TDE, MrSQM, and DrCIF, and achieves broadly similar accuracy to {\rocket}, InceptionTime, and TS-CHIEF, but is less accurate than HC2.  On average, {\multirocket}+{\hydra} is not significantly less accurate than HC2.

\begin{figure}
\centering
\includegraphics[width=1.0\linewidth]{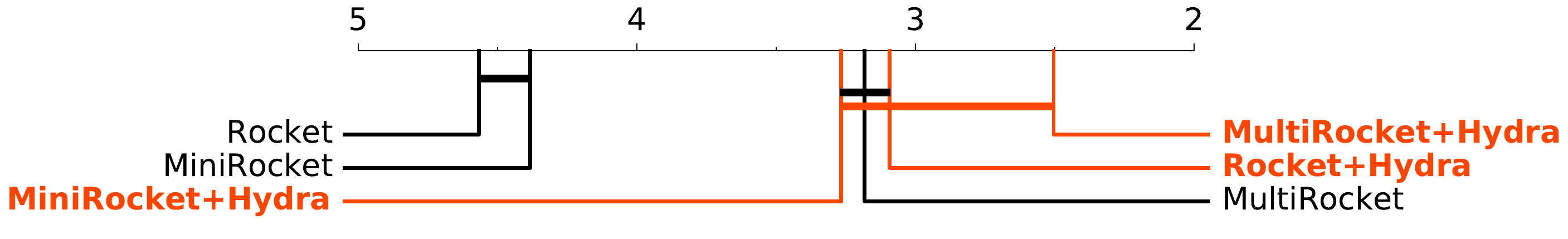}
\caption{Mean rank of {\rocket} and its variants, with and without {\hydra}, over the same 30 resamples of 112 datasets.}
\label{fig-rank-ucr-rockets}
\end{figure}

Figure \ref{fig-rank-ucr-rockets} shows the mean rank of {\rocket} and its variants, with and without {\hydra}.  (The pairwise differences between {\multirocket}, {\rocket}+{\hydra}, and {\minirocket}+{\hydra} are not statistically significant.  Similarly, the pairwise differences between {\rocket}+{\hydra}, {\minirocket}+{\hydra}, and {\multirocket}+{\hydra} are not statistically significant, as indicated by the orange line joining the relevant methods.)  The addition of {\hydra} significantly improves the accuracy of both {\rocket} and {\minirocket}.  {\rocket}+{\hydra} and {\minirocket}+{\hydra} both achieve similar accuracy to {\multirocket}.  This is consistent with results for a number of larger datasets: Section \ref{subsection-experiments-big}.  While {\multirocket}+{\hydra} is, on average, more accurate than {\multirocket}, this is largely an artefact of the large pairwise advantage of {\multirocket}+{\hydra} over {\multirocket}.  The actual differences in accuracy are mostly very small.  This is not necessarily surprising, as {\multirocket} is already one of the two most accurate methods for time series classification on the datasets in UCR archive, and already produces a large number of features.

\begin{figure}
\centering
\includegraphics[width=1.0\linewidth]{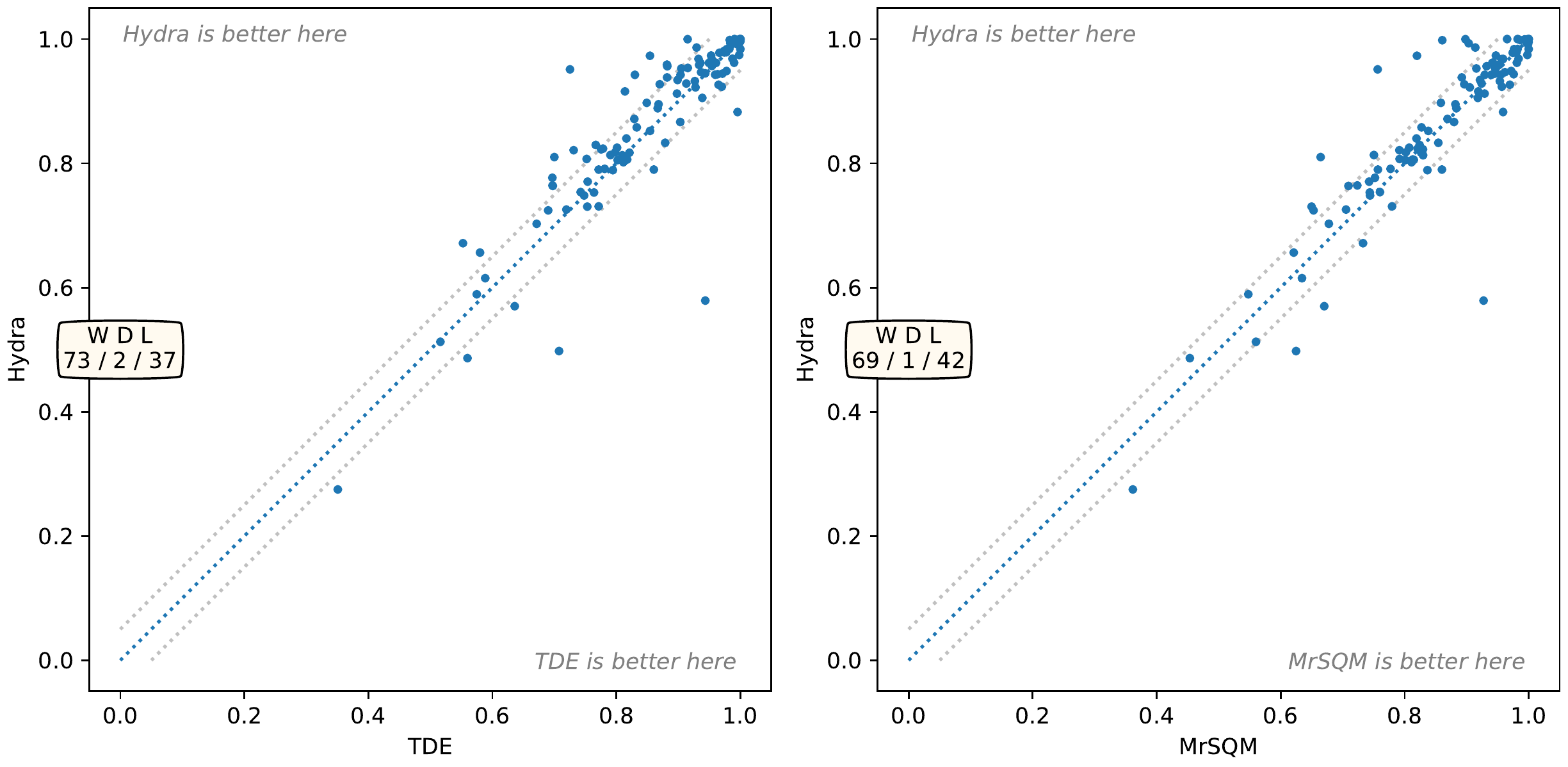}
\caption{Pairwise accuracy of {\hydra} vs TDE (left) and MrSQM (right).}
\label{fig-pairwise-tde-mrsqm}
\end{figure}

Figure \ref{fig-pairwise-tde-mrsqm} shows the pairwise accuracy of {\hydra} versus the two most accurate existing dictionary methods for time series classification: TDE (left) and MrSQM (right).  {\hydra} is more accurate than TDE on 73 datasets, and less accurate on 37.  Similarly, {\hydra} is more accurate than MrSQM on 69 datasets, and less accurate on 42.

\begin{figure}
\centering
\includegraphics[width=1.0\linewidth]{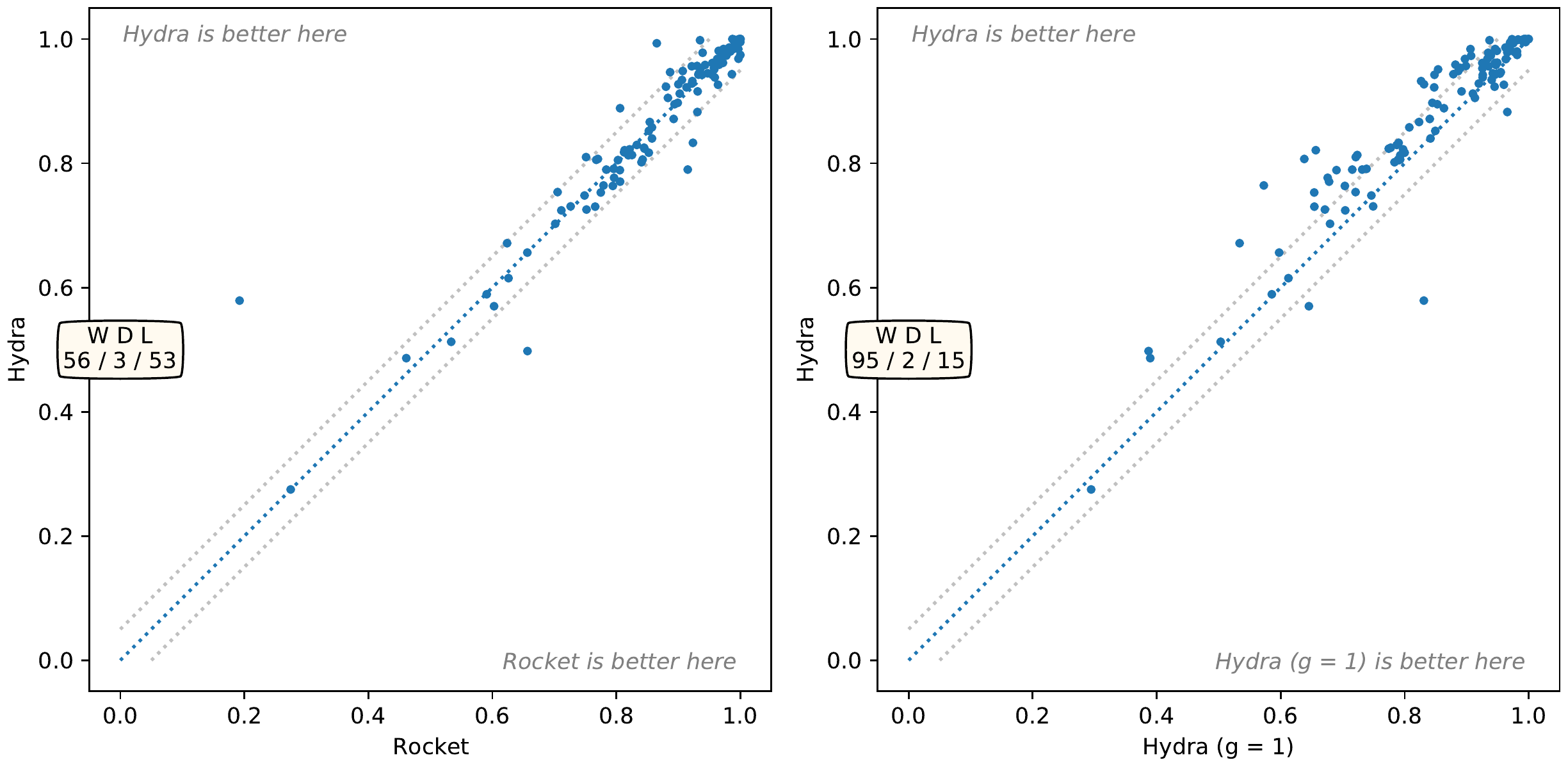}
\caption{Pairwise accuracy of {\hydra} vs {\rocket} (left) and {\hydra} ($g = 1$) (right).}
\label{fig-pairwise-rocket-g1}
\end{figure}

Figure \ref{fig-pairwise-rocket-g1} shows the pairwise accuracy of {\hydra} versus {\rocket} (left), and an alternative configuration of {\hydra} using a single group (right).  (This is the most dictionary-like variant of {\hydra}, using a single group and hard counting.)  {\hydra} is more accurate than {\rocket} on 56 datasets, and less accurate on 53.  The default variant of {\hydra} ($k=8/g=64$) is more accurate than the most dictionary-like variant of {\hydra} ($k=512/g=1$) on 95 datasets, and less accurate on 15.

\begin{figure}
\centering
\includegraphics[width=1.0\linewidth]{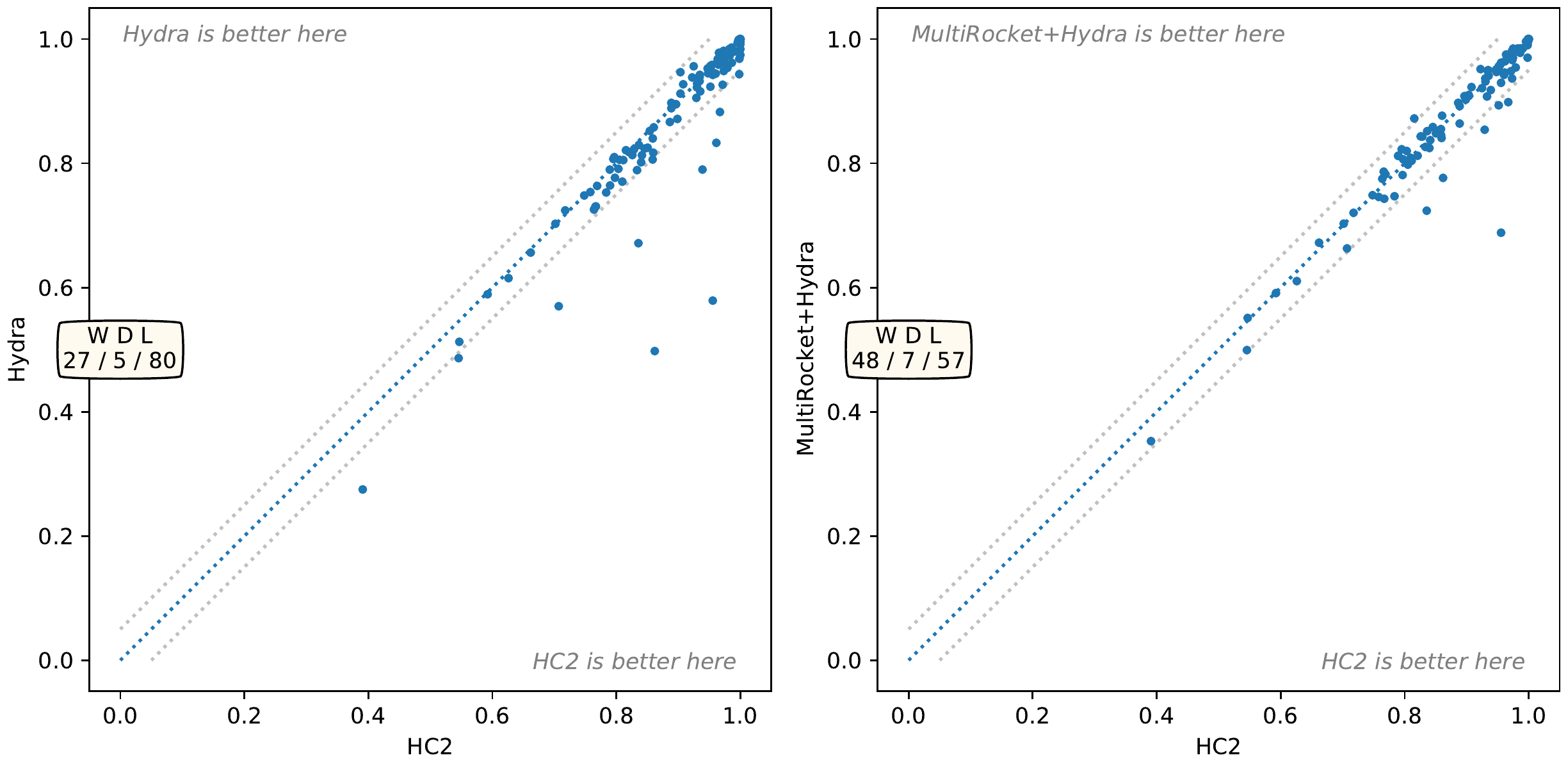}
\caption{Pairwise accuracy of {\hydra} (left) and {\multirocket}+{\hydra} (right) vs HC2.}
\label{fig-pairwise-hc2}
\end{figure}

Figure \ref{fig-pairwise-hc2} shows the pairwise accuracy of {\hydra} (left) and {\multirocket}+{\hydra} (right) versus HC2.  {\hydra} is more accurate than HC2 on 27 datasets and less accurate on 80.  {\multirocket}+{\hydra} is more accurate than HC2 on 48 datasets and less accurate on 57 (compared to 45/62 for {\minirocket}+{\hydra} and 44/62 for {\rocket}+{\hydra}).

While the pairwise comparison between {\hydra} and HC2 is heavily in favour of HC2, the actual differences in accuracy are mostly relatively small.  In contrast, the difference in computational cost is enormous.  The magnitude of the pairwise difference between {\hydra} and HC2 is: less than $5\%$ for $90\%$ of datasets ($94\%$ for {\multirocket}+{\hydra}); less than $2.5\%$ for $72\%$ of datasets ($85\%$ for {\multirocket}+{\hydra}); and less than $1\%$ for $53\%$ of datasets ($57\%$ for {\multirocket}+{\hydra}).  However {\hydra}, like {\rocket} and its variants, requires only a small fraction of the compute time of other methods of comparable accuracy: less than $1\%$ of the compute time of TDE, and less than $0.2\%$ of the compute time of HC2.  In other words, the advantages of HC2 over {\hydra} in terms of accuracy come at a $500 \times$ computational cost.

\subsection{Larger Datasets} \label{subsection-experiments-big}

We demonstrate the accuracy and scalability of {\hydra} on the three largest datasets in the UCR archive, namely \textit{MosquitoSound} ($139{,}780$ training examples, each of length $3{,}750$), \textit{InsectSound} ($25{,}000$ training examples, each of length $600$), and \textit{FruitFlies} ($17{,}259$ training examples, each of length $5{,}000$).  These newer additions to the archive are separate from the 112 datasets referred to in earlier experiments, and are significantly larger than the other datasets in the archive.

\begin{table}
  \centering
  \caption{Accuracies of \textsc{Mini}/\textsc{Multi}/{\rocket}, and {\hydra}, on the three largest datasets.}
  \label{table-big-1}
  \begin{tabular}{ccccc}
    \toprule
    {} &  {\rocket} &  \textsc{Mini} &  \textsc{Multi} &  {\hydra} \\
    \midrule
    \textit{FruitFlies}    &     0.9460 &         0.9555 &          0.9688 &    0.9674 \\
    \textit{InsectSound}   &     0.7823 &         0.7667 &          0.8038 &    0.7914 \\
    \textit{MosquitoSound} &     0.8253 &         0.8126 &          0.8622 &    0.8337 \\
    \bottomrule
  \end{tabular}
\end{table}

\begin{table}
  \centering
  \caption{Accuracies of \textsc{Mini}/\textsc{Multi}/{\rocket}+{\hydra} on the three largest datasets.}
  \label{table-big-2}
  \begin{tabular}{cccc}
    \toprule
    {} &  {\rocket}\textsuperscript{+} &  \textsc{Mini}\textsuperscript{+} &  \textsc{Multi}\textsuperscript{+} \\
    \midrule
    \textit{FruitFlies}    &     0.9676 &         0.9702 &          0.9728 \\
    \textit{InsectSound}   &     0.8050 &         0.7921 &          0.8118 \\
    \textit{MosquitoSound} &     0.8526 &         0.8485 &          0.8724 \\
    \bottomrule
  \end{tabular}
\end{table}

For this purpose, we integrate {\hydra} and the {\rocket} family of methods with logistic regression trained using Adam per \citet{dempster_etal_2021}.  Full training details are provided in Appendix \ref{appendix-big}.

The results for {\rocket}, {\minirocket}, {\multirocket}, and {\hydra} are shown in Table \ref{table-big-1}.  The results for the combination of {\hydra} with {\rocket}, {\minirocket}, and {\multirocket} are shown in Table \ref{table-big-2}.  {\hydra} is combined with the other methods by simply concatenating the features generated by each transform.

{\hydra} is more accurate than either {\rocket} or {\minirocket} on these datasets.  The addition of {\hydra} meaningfully improves the accuracies of {\rocket}, {\minirocket}, and {\multirocket}.  The combination of {\multirocket}+{\hydra} produces the highest accuracy on all three datasets.

Training times are shown in Tables \ref{table-big-timing-1} \& \ref{table-big-timing-2}, Appendix \ref{appendix-big}.  As expected, {\minirocket} is the fastest method, and is almost an order of magnitude faster than any other method on the largest dataset.  {\hydra} is slower than {\multirocket}, but the combination of {\multirocket}+{\hydra} is still marginally faster than {\rocket} on all three datasets.  Note that the absolute timings are somewhat arbitrary, as all methods can be run using more or fewer cores, and all methods can also be adapted to run on GPU.

\subsection{Sensitivity Analysis} \label{subsection-experiments-sensitivity}

We explore the effect of key hyperparameters in terms of:
\begin{itemize}
  \item the number of groups ($g$), and the number of kernels per group ($k$);
  \item the way in which the kernels are counted;
  \item whether or not to include the first-order difference; and
  \item whether or not to `clip' the convolution output.
\end{itemize}

We conduct the sensitivity analysis using the subset of 40 `development' datasets (default training/test splits) from the UCR archive per \citet{dempster_etal_2020,dempster_etal_2021}: see Section \ref{section-method}.  Results are mean results over 10 runs.

\subsubsection{Number of Groups ($g$) \& Kernels per Group ($k$)}

\begin{figure}
\centering
\includegraphics[width=1.0\linewidth]{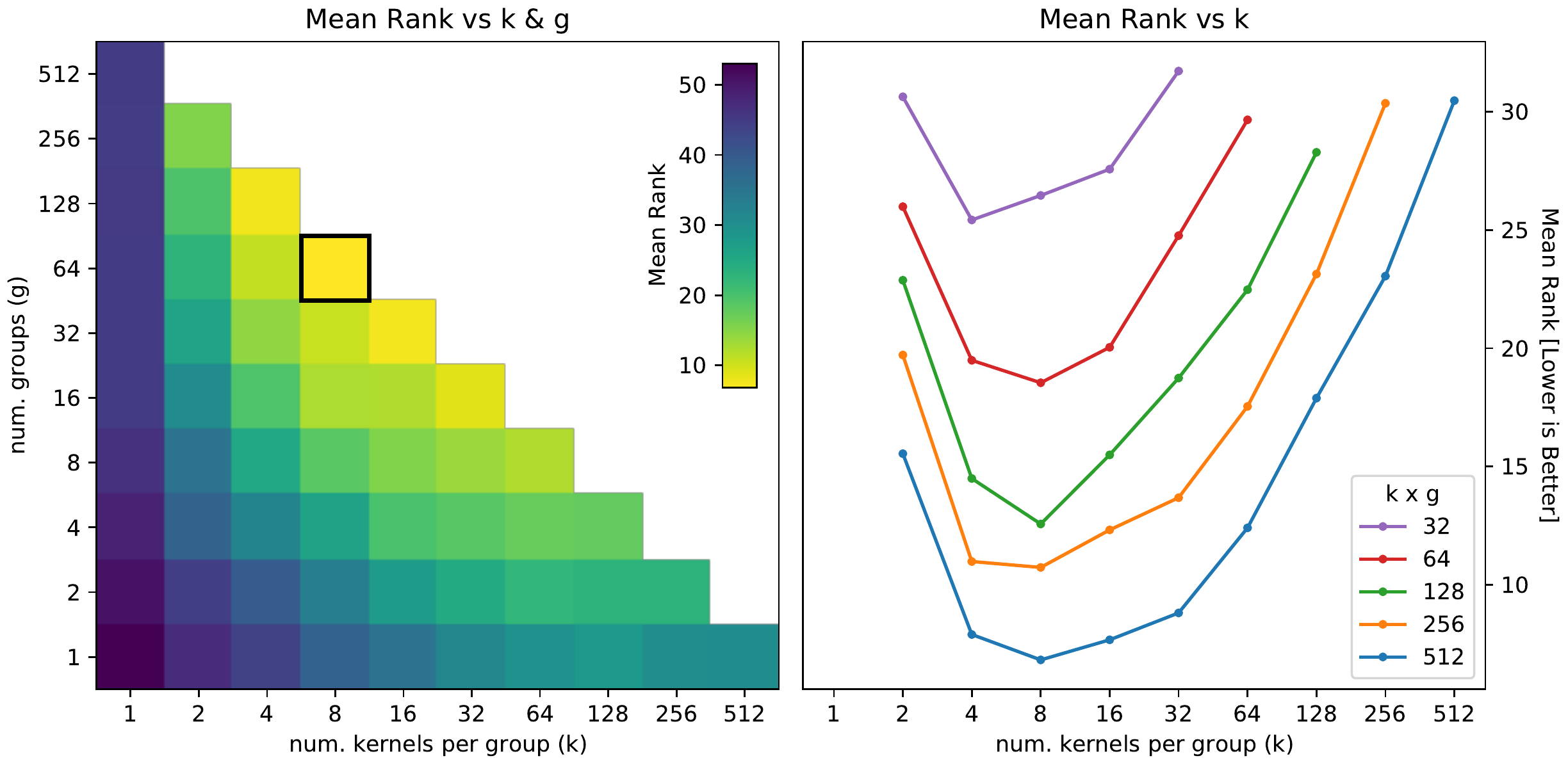}
\caption{Mean rank (accuracy) versus the number of kernels per group ($k$), and the number of groups ($g$).}
\label{fig-k-vs-g}
\end{figure}

Figure \ref{fig-k-vs-g} shows the relationship between accuracy and the number of groups ($g$), and kernels per group ($k$), for the most accurate variant of {\hydra} (counting both the maximum and minimum response, using both soft and hard counting, including the first-order difference, and no clipping: see Sections \ref{subsubsection-counting} \& \ref{subsubsection-clipping}).  Accuracy is represented by mean rank: lower mean rank corresponds to higher accuracy.  For each dataset, we rank the accuracy of each combination of $k$ and $g$, then take the mean rank over all datasets.

Figure \ref{fig-k-vs-g} (left) shows that 64 groups of 8 kernels ($k=8/g=64$) produces the highest accuracy.  Interestingly, this result is consistent across every hyperparameter configuration: counting the maximum response or both the maximum response and minimum response, soft, hard, or both soft and hard counting, with or without the first-order difference, and with or without clipping.  Additional results for the same model with and without the first-order difference, and with and without clipping, are shown in Appendix \ref{appendix-sensitivity}.

Figure \ref{fig-k-vs-g} (right) shows the same mean ranks versus $k$.  Each line represents a different total number of kernels.  This shows that, except for a very small total number of kernels, the optimal value of $k$ is approximately 8, suggesting that for any given `budget' ($k \times g$), it is more effective to increase the number of groups, keeping the number of kernels per group at approximately 8.

This also shows that, in general, adding more kernels increases accuracy.  For any number of kernels per group ($k$), a lager number of groups ($g$)---in other words, a larger total number of kernels ($k \times g$)---is more accurate.

The number of kernels represents a tradeoff between accuracy and computational efficiency.  While increasing the total number of kernels tends to increase accuracy, the magnitude of the improvement decreases as the number of kernels increases.  Although there is a clear difference in rank between $k \times g = 256$ and $k \times g = 512$, the actual differences in accuracy are small: see Figure \ref{fig-512-vs-256}, Appendix \ref{appendix-sensitivity}.  Increasing the total number of kernels in order to meaningfully increase accuracy beyond $k \times g = 512$ (e.g., $k \times g = 1024$) would involve considerable additional computational burden for little practical gain.

Additionally, limiting the total number of kernels per dilation ($k \times g$) to $512$ means that the total number of features per dataset is less than $10{,}000$ (at least for the 40 `development' datasets) or, in other words, roughly commensurate with {\minirocket}.  For these reasons, and given that {\hydra} is already approximately as accurate as {\rocket}, we set $k \times g = 512$ as the maximum number of kernels per dilation.

Different values of $k$ represent different levels of approximation in the kernels: see Section \ref{subsubsection-approximation}.  Figure \ref{fig-k-vs-g} shows that a relatively high degree of approximation is desirable, broadly consistent with the high level of approximation typical of dictionary methods.  The most accurate variants of {\hydra} are high-$g$/low-$k$ models, i.e., using a large number of small groups rather than a small number of large groups.  This implies that it is preferable to combine the information from many approximate matches.  A large number of small groups also has the benefit of low variability: see Figure \ref{fig-k-vs-g-std}, Appendix \ref{appendix-sensitivity}.

We also observe that, as $k$ increases, there is increasing sparsity, i.e., an increasing proportion of kernels which are never counted, although there is large variability between datasets: see Figure \ref{fig-sparsity}, Appendix \ref{appendix-sensitivity}.  For the optimal combination of $k$ and $g$ (i.e., $k = 8/g=64$), {\hydra} produces dense features.

\subsubsection{Method of Counting and First-Order Difference} \label{subsubsection-counting}

\begin{figure}
\centering
\includegraphics[width=1.0\linewidth]{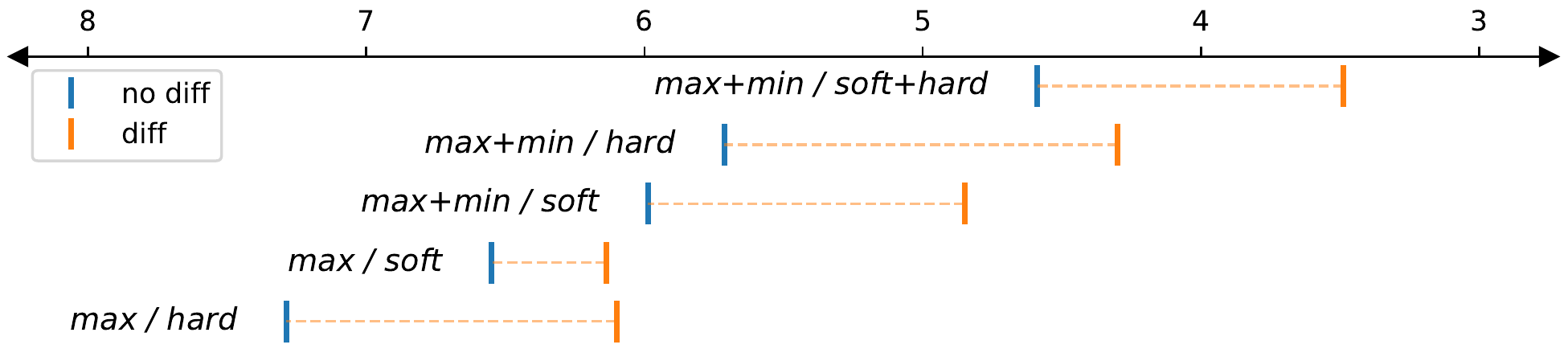}
\caption{Mean rank of different variants of {\hydra}, with and without first-order difference.}
\label{fig-rank-diff}
\end{figure}

Figure \ref{fig-rank-diff} shows the mean rank of different variants of {\hydra}, with and without the first-order difference (for the most accurate combination of $k/g$, being $k=8/g=64$ in all cases, and in all cases not using clipping).  Figure \ref{fig-rank-diff} shows that counting both the maximum and minimum responses is more accurate than only counting the maximum response, that using both soft and hard counting is more accurate than using either soft or hard counting alone, and that including the first-order difference always improves accuracy.

\subsubsection{Clipping} \label{subsubsection-clipping}

\begin{figure}
\centering
\includegraphics[width=1.0\linewidth]{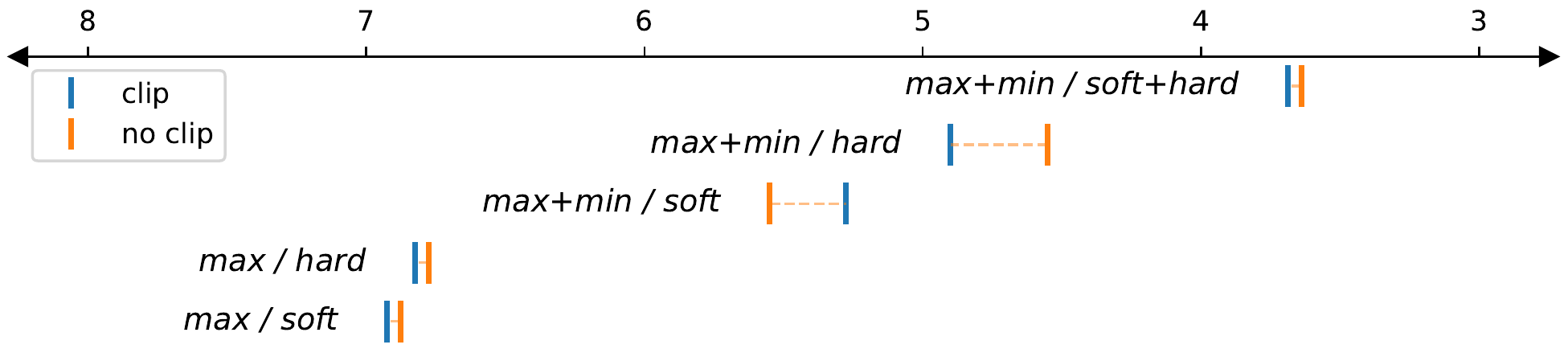}
\caption{Mean rank of different variants of {\hydra}, with and without clipping.}
\label{fig-rank-clip}
\end{figure}

Figure \ref{fig-rank-clip} shows the mean rank of different variants of {\hydra}, with and without clipping (for the most accurate combination of $k/g$, being $k=8/g=64$ in all cases, and in all cases including the first-order difference).  Figure \ref{fig-rank-clip} shows that clipping has very little effect on accuracy for optimal values of $k/g$.

\section{Conclusion}

We demonstrate a simple connection between dictionary methods for time series classification, which extract and count symbolic patterns in time series, and methods based on transforming the input time series using convolutional kernels, namely {\rocket}.  This provides the basis for fast and accurate dictionary-like classification using convolutional kernels.  We present {\hydra}, a simple, fast, and highly accurate dictionary method for time series classification, incorporating aspects of both {\rocket} and conventional dictionary methods.  {\hydra} demonstrates the advantages of performing dictionary-like classification with convolutional kernels and random patterns.  {\hydra} is faster and more accurate than the most accurate existing dictionary methods, and consistently improves the accuracy of {\rocket} and its variants when used in combination with these methods.  In future work we plan to explore deterministic variants of {\hydra}, more sophisticated approaches to combining {\hydra} with other methods, and the application of {\hydra} to multivariate time series.

\bmhead{Acknowledgements}

This material is based on work supported by an Australian Government Research Training Program Scholarship and the Australian Research Council under award DP210100072.  The authors would like to thank Professor Eamonn Keogh and all the people who have contributed to the UCR time series classification archive.  Figures showing mean ranks were produced using code from \citet{ismailfawaz_etal_2019}.

\bibliography{references}

\clearpage

\appendix

\section{Additional Results} \label{appendix-sensitivity}

\begin{figure}[h]
\centering
\includegraphics[width=0.5\linewidth]{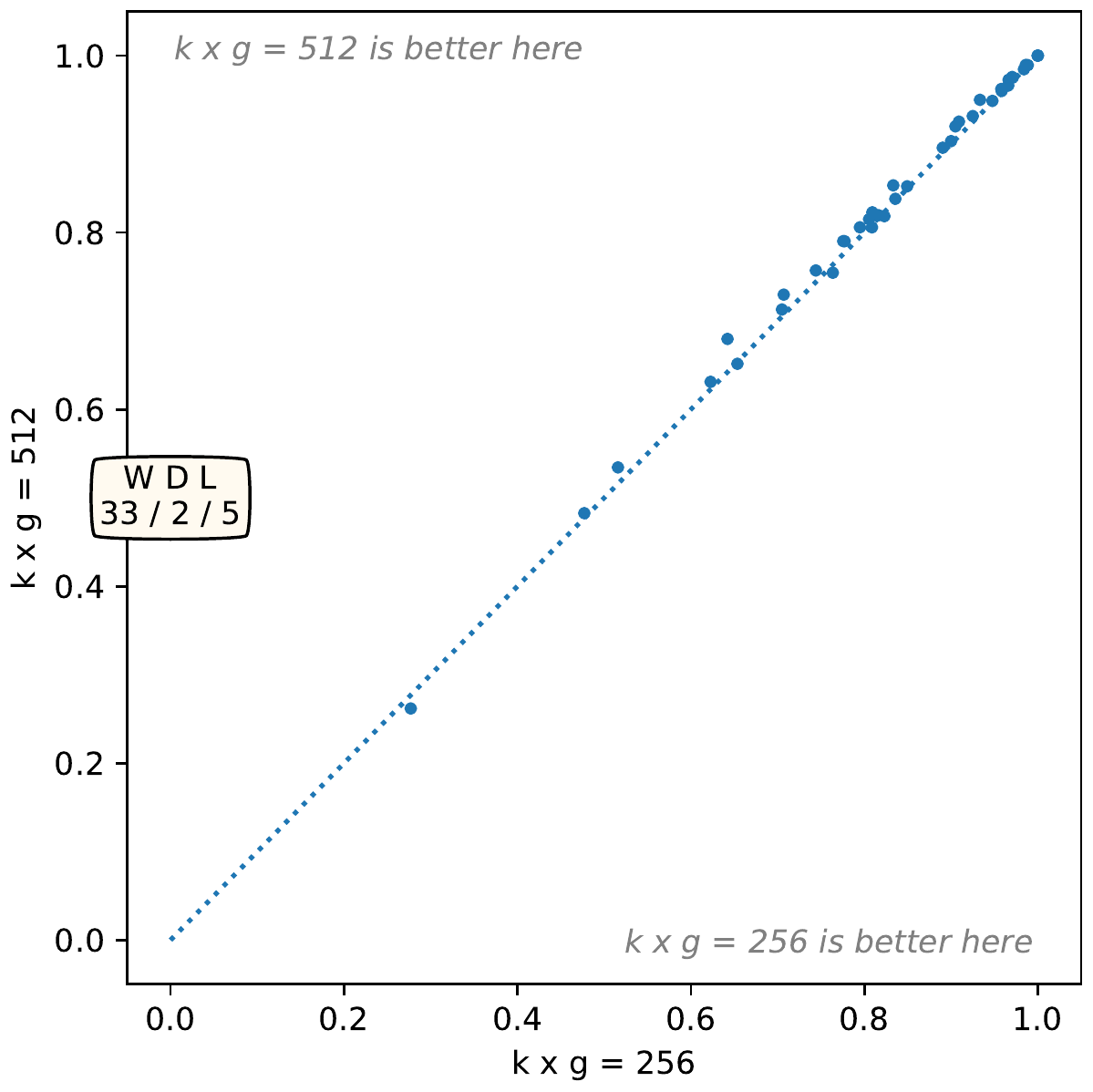}
\caption{Pairwise accuracy of $k \times g = 512$ vs $k \times g = 256$.}
\label{fig-512-vs-256}
\end{figure}

\vspace{20mm}

\begin{figure}[h]
\centering
\includegraphics[width=0.5\linewidth]{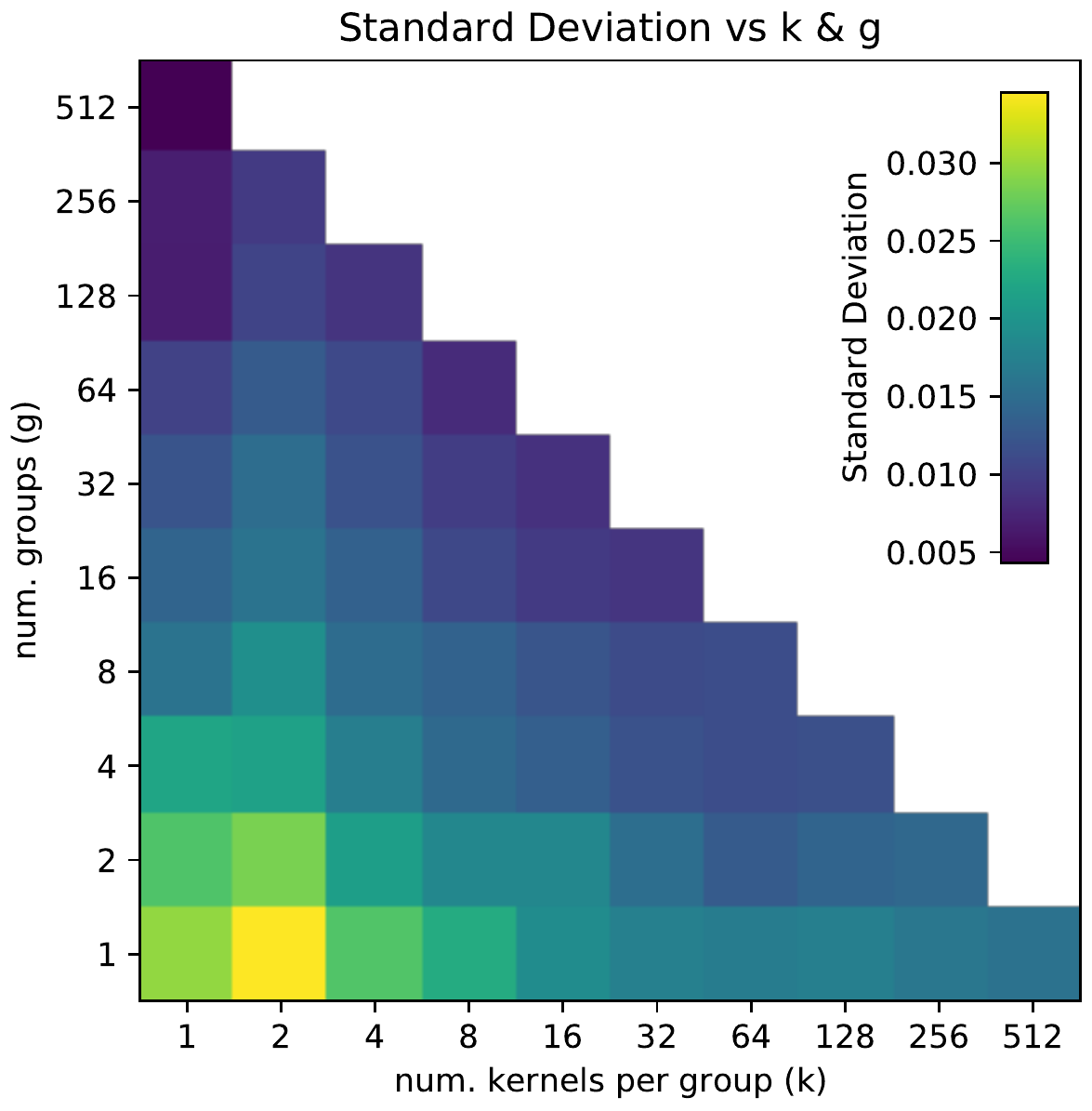}
\caption{Standard deviation (accuracy) vs $k$ \& $g$.}
\label{fig-k-vs-g-std}
\end{figure}

\begin{figure}[h]
\centering
\includegraphics[width=0.5\linewidth]{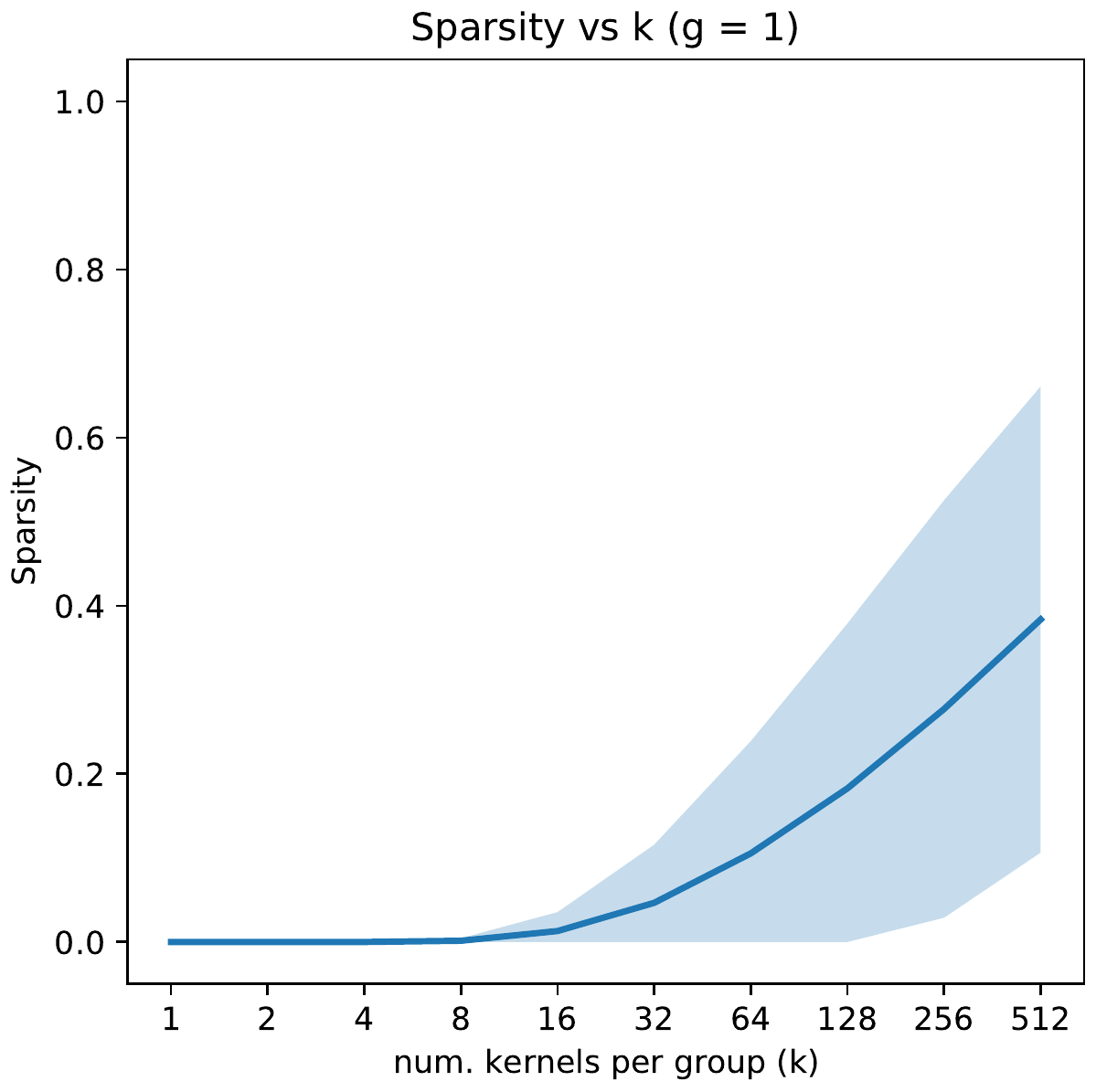}
\caption{Sparsity vs $k$ ($g=1$).}
\label{fig-sparsity}
\end{figure}

\begin{figure}[h]
\centering
\includegraphics[width=1.0\linewidth]{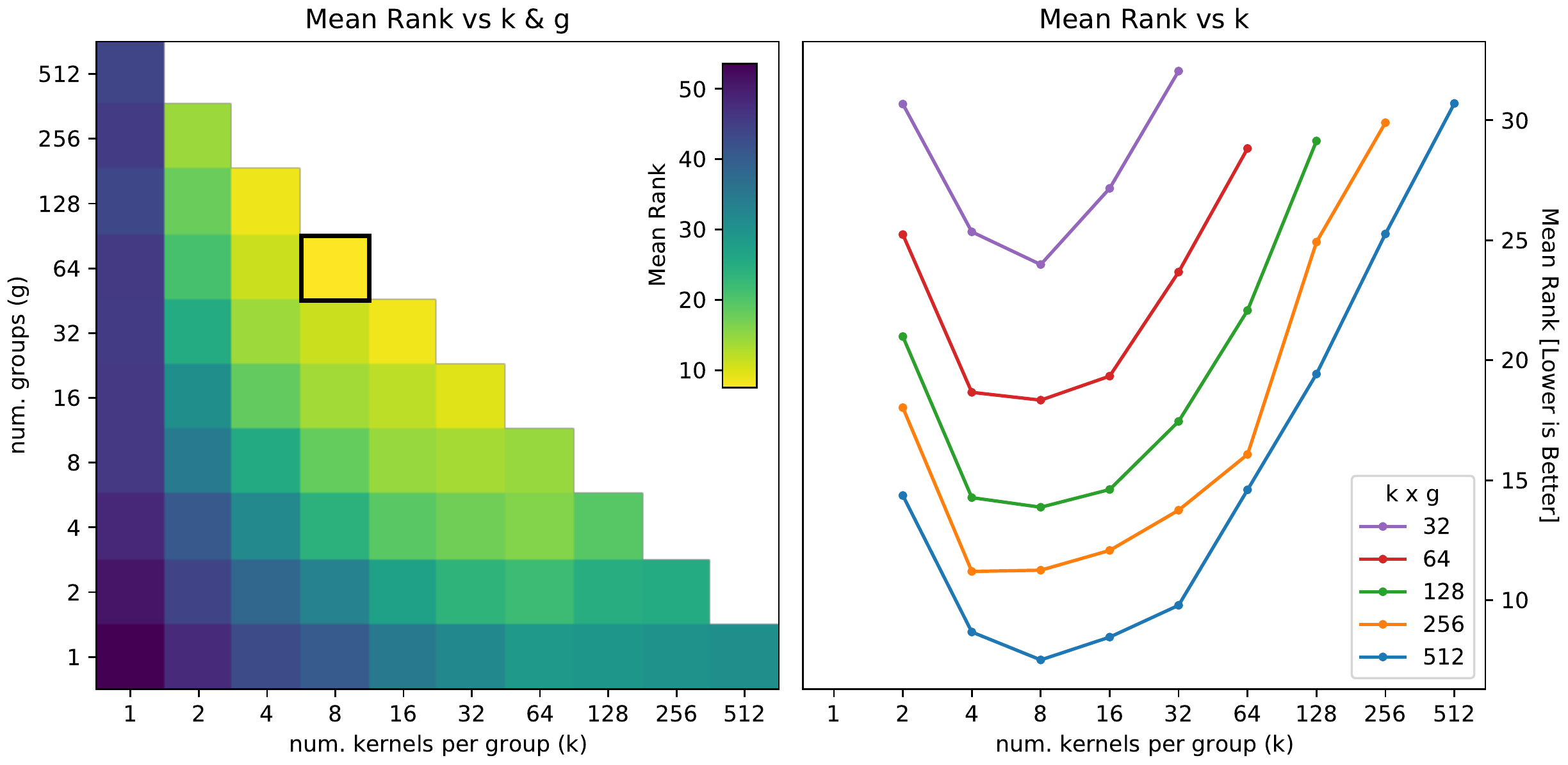}
\caption{Mean rank vs $k$ \& $g$ (\textit{max+min/soft+hard/no diff/no clip}).}
\label{fig-kg-extra-1}
\end{figure}

\begin{figure}[h]
\centering
\includegraphics[width=1.0\linewidth]{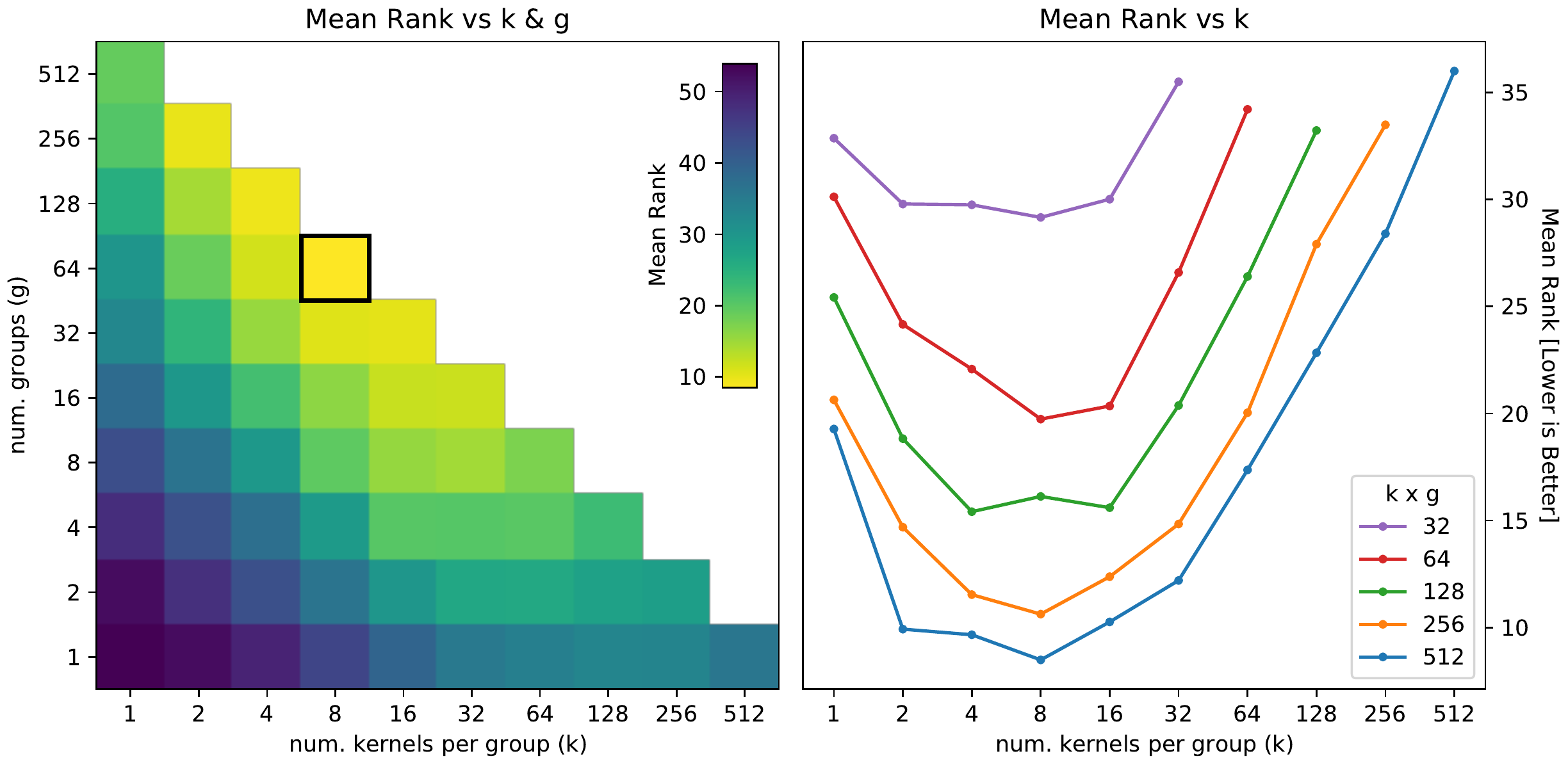}
\caption{Mean rank vs $k$ \& $g$ (\textit{max+min/soft+hard/no diff/clip}).}
\label{fig-kg-extra-2}
\end{figure}

\begin{figure}[h]
\centering
\includegraphics[width=1.0\linewidth]{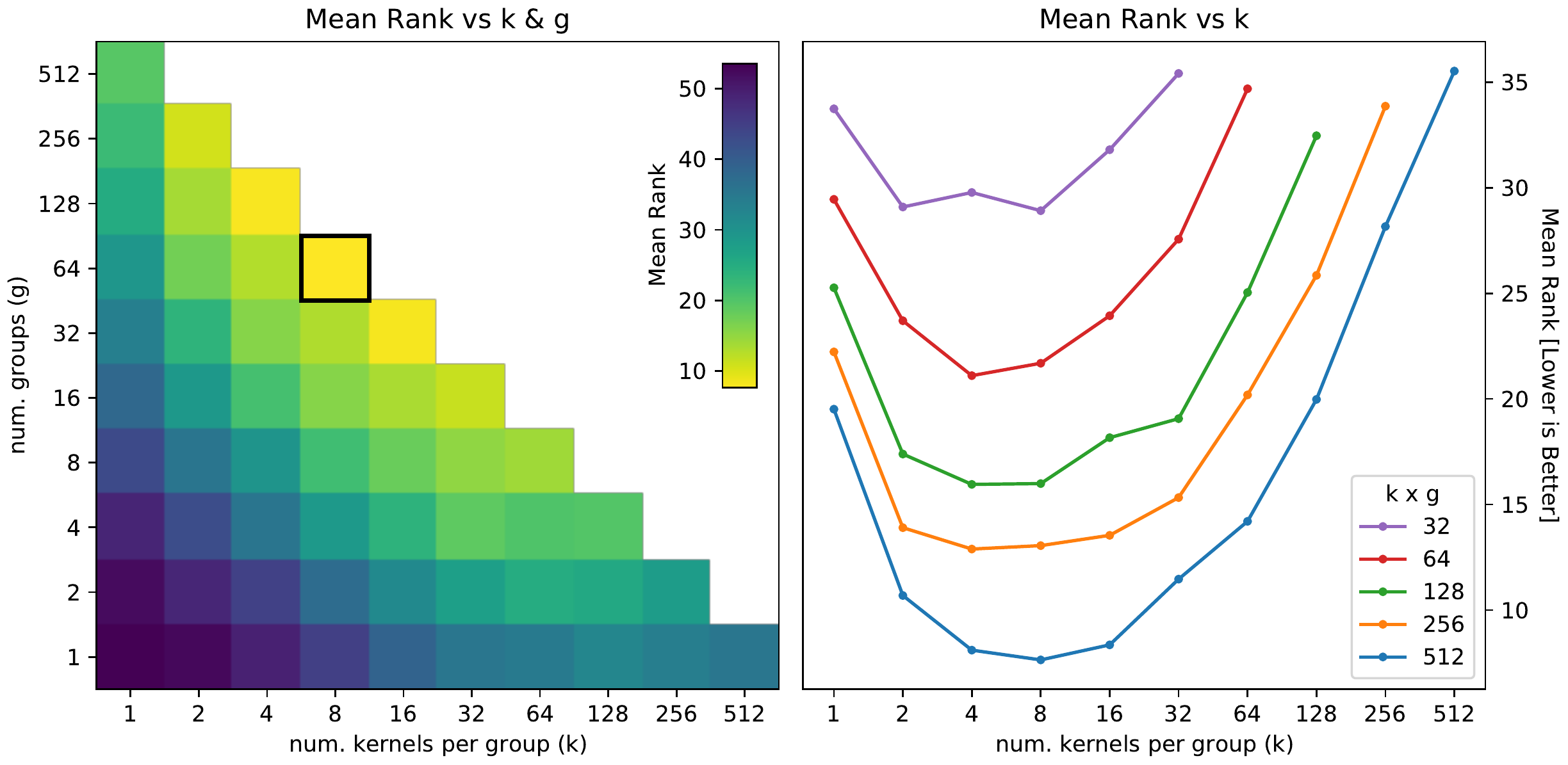}
\caption{Mean rank vs $k$ \& $g$ (\textit{max+min/soft+hard/diff/clip}).}
\label{fig-kg-extra-3}
\end{figure}

\clearpage

\section{Larger Datasets} \label{appendix-big}

\subsection{Training Time}

\begin{table}[h]
  \centering
  \caption{Training times for \textsc{Mini}/\textsc{Multi}/{\rocket} and {\hydra} on the three largest datasets.}
  \label{table-big-timing-1}
  \begin{tabular}{crrrr}
    \toprule
    {} & {\rocket} & \textsc{Mini} & \textsc{Multi} &   {\hydra} \\
    \midrule
    \textit{FruitFlies} & 20m 14s & 56s & 7m 16s & 12m 12s \\
    \textit{InsectSound} & 3m 28s & 14s & 1m 18s & 1m 45s \\
    \textit{MosquitoSound} & 2h 2m 7s & 4m 8s & 38m 47s &  1h 4m 12s \\
    \bottomrule
  \end{tabular}
\end{table}

\begin{table}[h]
  \centering
  \caption{Training times for \textsc{Mini}/\textsc{Multi}/{\rocket}+{\hydra} on the three largest datasets.}
  \label{table-big-timing-2}
  \begin{tabular}{crrr}
    \toprule
    {} & {\rocket}\textsuperscript{+} & \textsc{Mini}\textsuperscript{+} & \textsc{Multi}\textsuperscript{+} \\
    \midrule
    \textit{FruitFlies} & 32m 1s & 12m 16s & 19m 3s \\
    \textit{InsectSound} & 5m 1s & 2m 12s & 2m 52s \\
    \textit{MosquitoSound} & 2h 58m 35s & 1h 9m 8s & 1h 43m 4s \\
    \bottomrule
  \end{tabular}
\end{table}

\subsection{Training Details}

For each method, the transform is performed in larger batches of $4{,}096$ examples, further divided into minibatches for training.  For {\minirocket}  and {\multirocket}, we fit the bias values using the first such batch of $4{,}096$ examples.  We cache the transformed features in order to avoid unnecessarily repeating the transform when training for multiple epochs.  We use the Adam optimizer \citep{kingma_and_ba_2015}, and we use the same hyperparameters for all methods and datasets: a validation set of $2{,}048$ training examples, a minibatch size of $256$, and an initial learning rate of $10^{-4}$.  The learning rate is halved if validation loss does not improve after 50 updates, and training is stopped if validation loss does not improve after 100 updates (but only after the first epoch).

We run the experiments on the same cluster referred to in Section \ref{section-introduction}, performing five separate runs per dataset for each method (results are mean results over those five runs), and using eight cores per dataset per run.

\end{document}